\RequirePackage{lineno}
\documentclass[preprint,12pt,authoryear]{elsarticle}

\usepackage{amssymb}

\usepackage{amsmath}

\usepackage{booktabs}
\usepackage{makecell}
\usepackage{multirow}
\usepackage{graphicx}

\usepackage{algorithm2e}

\begin{document}

\begin{frontmatter}

\makeatletter
\def\ps@pprintTitle{%
  \let\@oddfoot\@empty
  \let\@evenfoot\@oddfoot
}



\title{Improving Interpretability of Deep Active Learning for Flood Inundation Mapping Through Class Ambiguity Indices Using Multi-spectral Satellite Imagery}

\author{Hyunho Lee\fnref{asu}}
\author{Wenwen Li\corref{cor}\fnref{asu}}
\ead{wenwen@asu.edu}
\affiliation[asu]{organization={School of Geographical Sciences and Urban Planning, Arizona State University},
           city={Tempe},
           postcode={85287-5302}, 
           state={AZ},
           country={USA}}
\cortext[cor]{Corresponding author.}




\begin{abstract}
Flood inundation mapping is a critical task for responding to the increasing risk of flooding linked to global warming. Significant advancements of deep learning in recent years have triggered its extensive applications, including flood inundation mapping. To cope with the time-consuming and labor-intensive data labeling process in supervised learning, deep active learning strategies are one of the feasible approaches. However, there remains limited exploration into the interpretability of how deep active learning strategies operate, with a specific focus on flood inundation mapping in the field of remote sensing. In this study, we introduce a novel framework of Interpretable Deep Active Learning for Flood inundation Mapping (IDAL-FIM), specifically in terms of class ambiguity of multi-spectral satellite images. In the experiments, we utilize Sen1Floods11 dataset, and adopt U-Net with MC-dropout. In addition, we employ five acquisition functions, which are the random, K-means, BALD, entropy, and margin acquisition functions. Based on the experimental results, we demonstrate that two proposed class ambiguity indices are effective variables to interpret the deep active learning by establishing statistically significant correlation with the predictive uncertainty of the deep learning model at the tile level. Then, we illustrate the behaviors of deep active learning through visualizing two-dimensional density plots and providing interpretations regarding the operation of deep active learning, in flood inundation mapping.
\end{abstract}




\end{frontmatter}

\section{Introduction}
\label{Introduction}

Flood inundation mapping, which determines the extent of the flooded area including depth, velocity and uncertainty \citep{bentivoglio2022deep, merwade2008uncertainty, horritt2006methodology}, is increasingly important due to the intensification of extreme precipitation worldwide. 
This intensification is anticipated due to global warming. Rising Earth's average temperatures lead to higher water vapor concentrations in the atmosphere, consequently contributing to more extreme precipitation occurrences \citep{tabari2020climate}. 
Significantly, the extreme values, representing the 90th percentile value of precipitation duration for each year globally, of long-duration flood events have exceeded 30 days in the recent decade, whereas they were less than 20 days in the 1980s and 1990s \citep{najibi2018recent}. In addition, between 2000 and 2018, an estimated 255-290 million people were directly affected by floods in areas observed by satellites \citep{tellman2021satellite}. Therefore, to respond to the risks posed by floods, flood inundation mapping plays a fundamental role in near real-time monitoring, damage assessment, post-flood evacuation, and protection planning \citep{bentivoglio2022deep, iqbal2021computer}. 

In the past decade, notable advancements have been made in deep learning, particularly with the introduction of Convolutional Neural Networks \citep[CNNs;][]{jia2014caffe}. These advances have enabled automated and data-driven analysis of large imagery, and they have also triggered extensive applications of deep learning in environmental monitoring using remote sensing imagery \citep{li2022geoai, li2024segment}.
Furthermore, this research trend, coupled with advances in Earth observation data and high-performance computing, has led to the emergence of Geospatial Artificial Intelligence \citep[GeoAI;][]{li2020geoai}, an interdisciplinary research area that applies and extends AI for geospatial problem solving. 

Flood inundation mapping with remote sensing images, an important application of GeoAI, is primarily focused on identifying flooded areas from the given satellite images using deep learning models. Therefore, research in this area has predominantly centered on semantic segmentation which partitions an image into distinct regions corresponding to predefined classes. Previous research have shown that deep learning models, such as Fully Convolutional Neural Network \citep[FCN;][]{long2015fully}, U-Net \citep{ronneberger2015u}, DeepLabV3+ \citep{chen2018encoder}, HRNet \citep{wang2020deep}, outperformed traditional methods including rule- and threshold-based approaches in flood mapping \citep{dong2021monitoring, helleis2022sentinel}. Additionally, new learning strategies, such as dilated convolution \citep{yu2015multi, yu2017dilated}, were integrated into deep learning-based segmentation models to further improve the models’ predictive performance \citep{nogueira2018exploiting, wang2022fwenet}. Recently, new geospatial foundation models such as Prithvi \citep{li2023assessment} were applied to flood mapping to assess their generalizability.

Despite these advances in AI model architecture for flood inundation mapping, the time-consuming and labor-intensive data annotation process remains a bottleneck in enabling supervised learning \citep{buscombe2022human, beluch2018power, takezoe2023deep}. The data labeling process can be divided into two stages: (1) selecting or sampling image tiles, and (2) labeling the selected tiles in satellite images. Active learning, which aims to identify a small yet highly informative set of data points for machine learning, is an effective approach to address the selection of data samples to reduce labeling cost \citep{settles2009active, cohn1996active}. In particular, active learning for deep learning models is referred to as deep active learning \citep{takezoe2023deep}. In the field of remote sensing, research has applied deep active learning to satellite image segmentation and change detection \citep{ruuvzivcka2020deep, li2022suggestive}. However, there remains limited exploration in interpreting how deep active learning operates, especially within the context of flood inundation mapping. This paper aims to bridge the knowledge gap by interpreting deep active learning in flood inundation mapping, with a specific focus on class ambiguity extracted from input satellite images.
\vspace{6pt}

The main contributions of this study are:
\vspace{6pt}
\begin{itemize}
  \item[] \noindent (1)  We introduce a novel framework of Interpretable Deep Active Learning for Flood inundation Mapping (IDAL-FIM) to enhance the interpretability of deep active learning operations. \vspace{3pt}
  \item[] \noindent (2) We demonstrate that the correlation between the two proposed class ambiguity indices (boundary pixel ratio and Mahalanobis distance for flood-segmentation) and predictive uncertainty of the deep learning model are statistically significant at the tile level. This finding allows us to interpret the behavior of deep active learning using the proposed indices. \vspace{3pt}
  \item[] \noindent (3)  We illustrate that the behaviors of deep active learning can be visually interpreted through two-dimensional density plots, which show the distribution patterns of selected data points to be labeled in the IDAL-FIM framework.
\end{itemize}

To achieve this research goal, the paper is structured as follows: Section 2 provides a review of relevant literature; Section 3 describes the IDAL-FIM framework, acquisition functions, and proposes two class ambiguity indices; Section 4 explains the experimental setup; Section 5 presents the experimental results; Section 6 provides a discussion and interpretation about the behavior of deep active learning in the context of flood mapping. Finally, in Section 7, we conclude the work, discuss limitations and propose future research directions.  

\section{Literature review}
\label{sec:Literature-review}

\subsection{Multi-spectral Satellite Image Collection for Deep Learning-Based Flood Inundation Mapping}
\label{sec:Multi-spectral Satellite Image Collection for Deep Learning-Based Flood Inundation Mapping}

There are two distinct approaches used to collect flood-observed training data for deep learning in flood inundation mapping: (1) region-specific satellite image collection and (2) global satellite image collection. The collection of multi-spectral satellite images which were captured during flood events in a specific region is only feasible when a sufficient amount of data can be acquired. Therefore, this approach is applicable to study areas that experience recurrent flood damage over the years and cover a relatively extensive geographical area. Examples of such study areas are the Yangtze River Basin and Lake Poyang in China, as well as the Atlantic coast of the southeastern United States, including nearby urban areas frequently affected by hurricanes \citep{peng2019patch,munoz2021local,wang2022fwenet,zhang2021flood}. 

On the other hand, flood events are infrequent hydrological phenomena; therefore, securing a sufficient amount of training samples in a specific region is almost infeasible, except for a few regions stated above. Instead, collecting training data containing flood events from diverse global locations has become a feasible solution. This has especially benefited from the availability of cloud platforms such as Google Earth Engine, NASA Earth Exchange, and Sentinel Hub which facilitate access to and processing of vast amounts of satellite imagery \citep{zhao2022overview}. During satellite image collection, researchers often acquire images across diverse climates, atmospheric conditions, and land settings to ensure the generalizability of deep learning models \citep{wieland2023semantic, shastry2023mapping, tellman2021satellite, bonafilia2020sen1floods11, wieland2019modular}.
However, to the best of our knowledge, there have been very few studies \citep{popien2021cost} investigating the impact of training data selection on the predictive performance of deep learning models in flood inundation mapping.

\subsection{Deep Active Learning}
\label{sec:Deep Active Learning}

\textbf{Active Learning} (AL) is designed to improve the performance of machine learning models by utilizing fewer training data \citep{settles2009active,cohn1996active}. The pool-based sampling scenario, employed in this study, is one of the typical scenarios of active learning, which assumes a large pool of unlabeled data points along with a small initial labeled data set. In each iteration, a model is trained using labeled data in a supervised learning manner. An acquisition function prioritizes informative data points and guides the selection of unlabeled data points from a pool. Unlabeled data selected through the acquisition function are labeled by human experts and then integrated into the existing training data. This iterative process is repeated, wherein the model is trained from scratch using the newly incorporated labeled data, until a specific level of model performance is reached \citep{beluch2018power,gal2017deep}. 

\textbf{Deep Active Learning} (DAL) combines the advantages of active learning, which effectively reduces labeling costs by selecting informative data points for model training, with a deep learning model, known for exceptional high-dimensional data processing and automatic feature extraction \citep{ren2021survey}. In DAL, the acquisition functions are mainly categorized into uncertainty-based and density-based acquisition functions \citep{takezoe2023deep,beluch2018power}. Both categories of acquisition functions are relying on specific assumptions to select informative data points. 

The uncertainty-based acquisition function evaluates the informativeness of unlabeled data under the assumption that data points with higher uncertainty provide more information for model training \citep{settles2009active}.  In the context of flood mapping, high uncertainty data points include satellite images capturing complex boundary patterns in flooded areas, as well as areas exhibiting spectral reflectance similar to flooded areas, such as non-flooded vegetated areas or cloud shadows. Deep learning models encounter more difficulty in classifying pixels in these satellite images into the correct classes. Therefore, by training on data points with high uncertainty, the model improves its ability to identify between classes in flood mapping, which can eventually enhance the model's performance \citep{takezoe2023deep}. 

Regarding uncertainty estimation in deep learning, recent research has pointed out that deep learning models often exhibit overconfidence in their predictions, especially when making misclassifications \citep{guo2017calibration}. To enhance the reliability of predictions, uncertainty estimation methods focus on calibrating the predictions instead of relying solely on a single prediction \citep{wang2023calibrating}. For this reason, prior studies on DAL \citep{ruuvzivcka2020deep,beluch2018power,gal2017deep} employed uncertainty estimation methods to measure more reliable predictions for relevant acquisition functions.

Particularly, in previous studies on the segmentation of remote sensing imagery, three uncertainty estimation methods were utilized to obtain uncertainty from deep learning models at the pixel-level: (1) Monte-Carlo dropout \citep[MC-dropout;][]{gal2016dropout}, (2) deep ensembles \citep{lakshminarayanan2017simple}, and (3) fully-Bayesian CNN \citep{labonte2019we}. 
The MC-dropout method derives uncertainty estimates by regarding dropout training in deep neural networks as an approximation of Bayesian inference within deep Gaussian processes \citep{gal2016dropout}. In practice, approximate Bayesian inference in deep learning models makes use of multiple inferences with different dropout masks. In remote sensing studies, the MC-dropout method was employed to enhance prediction performance and provide uncertainty estimation, in Land Use and Land Cover (LULC) tasks \citep{kampffmeyer2016semantic, dechesne2021bayesian}. On the other hand, deep ensembles utilize an ensemble of multiple deep learning models to estimate uncertainty, and the final output of deep ensembles is generally the averaged softmax vectors of each ensemble model \citep{beluch2018power}. In previous deep learning-based roads segmentation, deep ensembles were shown to outperform MC-dropout in pixel-level prediction, despite their significant computational cost \citep{haas2021uncertainty}. More recently, another study compared the reliability of uncertainty estimation methods between MC-dropout and fully-Bayesian CNN \citep{labonte2019we} in water segmentation \citep{hertel2023probabilistic}. The fully-Bayesian CNNs learn the distribution of the weight space instead of a single value. For implementation of the fully-Bayesian CNNs, the authors utilized the Bayesian Layers library \citep{tran2019bayesian} in TensorFlow Probability \citep{dillon2017tensorflow}. Their conclusion was that fully-Bayesian CNNs were more reliable than MC-dropout in estimating pixel-level predictive uncertainty \citep{hertel2023probabilistic}.
However, the implementation of full-Bayesian CNNs requires specific libraries and additional training time to determine weight distributions, compared to MC-dropout.

In contrast to uncertainty-based acquisition, density-based acquisition functions leverage the feature space of the input data \citep{takezoe2023deep}. This type of acquisition function is grounded in the assumption that data points maximizing the diversity of data features are informative \citep{xie2020deal}. However, while density-based acquisition functions have been primarily studied for classification tasks in computer vision, no relevant research on the segmentation of remote sensing images could be found.

\textbf{Uncertainty measures in deep active learning} are categorized into predictive uncertainty measures and model uncertainty measures. Predictive uncertainty is mainly estimated using a measure of entropy \citep{shannon1948mathematical} or margin \citep{scheffer2001active}, which is based on the class probability assigned to each pixel by the model. On the other hand, model uncertainty is commonly quantified by measuring the variance in predictions resulting from averaging over multiple models trained on consistent training data \citep{lakshminarayanan2017simple, gal2017deep}, such as Bayesian Active Learning by Disagreement \citep[BALD;][]{houlsby2011bayesian}.

\subsection{Deep Active Learning in the Field of Remote Sensing}
\label{sec:Deep Active Learning in the Field of Remote Sensing}

Flood inundation mapping mainly utilizes deep learning models for semantic segmentation to extract the distribution of water bodies along with detailed boundaries. However, most research on DAL in the field of remote sensing had focused on pixel classification, which focuses on predicting predefined classes by considering properties of a single pixel. Such work does not consider partitioning image scenes into semantically meaningful areas, known as the task of semantic segmentation. Even before deep active learning research, a substantial number of studies have investigated active learning for pixel classification, utilizing machine learning algorithms, including Support Vector Machine (SVM), Random Forest (RF), and Artificial Neural Network (ANN), across both multi-spectral and hyper-spectral imagery \citep{thoreau2022active,ruiz2013bayesian,stumpf2013active,pasolli2013svm,crawford2013active,li2011hyperspectral,tuia2011using,tuia2011survey,tuia2009active,rajan2008active,mitra2004segmentation}. Recently, there has been research on deep active learning for pixel classification using remote sensing imagery \citep{patel2023comprehensive,di2023active,cao2020hyperspectral}. In particular, various DAL studies have been conducted for pixel classification based on hyper-spectral satellite images. \citet{liu2016active} introduced a DAL scheme that utilizes the Deep Belief Network (DBN), and \citet{haut2018active} presented a DAL framework using CNNs with MC-dropout. In addition, \citet{lei2021active} proposed a DAL framework that includes an auxiliary light network, which is responsible for the uncertainty prediction of unlabeled samples. 

Unlike studies focused on pixel classification, research on DAL for semantic segmentation using remote sensing images is limited, with only a few studies in the field of remote sensing. \citet{ruuvzivcka2020deep} investigated deep active learning, employing deep ensembles and the Monte Carlo Batch Normalization (MCBN) method for change detection and map updating. The authors demonstrated that their proposed DAL framework not only identifies highly informative samples but also automatically balances classes in the training data within the specific number of samples, even in the presence of an extreme class imbalance in the pool of unlabeled data. In addition, \citet{li2022suggestive} proposed a DAL framework for building mapping to reduce the effort of data labeling. Their framework integrates two deep learning models, U-Net and DeepLabV3+, along with uncertainty-based acquisition functions. Furthermore, the authors utilized landscape metrics to provide a summary of the preliminary suggestions for data labeling. However, they only used landscape metrics to describe the characteristics of the selected data points in active learning, without quantifying the relationship between these indices and the operation of active learning. Thus far, based on our comprehensive review, there is a notable absence of studies interpreting the behavior of deep active learning in the field of remote sensing.

 \subsection{Uncertainty Propagation Theory and Uncertainty Descriptors in Remote Sensing}
\label{sec:Uncertainty Propagation and Uncertainty Descriptors in Remote Sensing}

Uncertainty propagation refers to quantifying how uncertainty in input or model parameter values affects the uncertainty of model predictions or computational procedure outputs \citep{wallach1998effect, lee2009comparative, crosetto2001uncertainty}. Research on uncertainty propagation in remote sensing has been conducted based on the recognition of its importance in ensuring the reliability and accuracy of high-level products essential for global change research and environmental management decision-making \citep{crosetto2001uncertainty}. In the case of the multi-spectral satellite imagery, the primary sources of uncertainty are sub-pixel mixing, spatial mis-registration, and sensor sampling bias \citep{bastin2002visualizing}.

In the previous study, \citet{zhang2019uncertainty} introduced two uncertainty descriptors, designed to quantitatively measure uncertainty when classifying pixels in remote sensing, based on the uncertainty propagation theory. The first descriptor, spatial distribution uncertainty, addresses the impact of adjacency effects within remote sensing images, while the second, semantic uncertainty, aims to quantify the considerable intra-class variations. The rationale behind the two descriptors is that, due to the spatial and spectral resolution limitations of the sensors, ambiguity arises at the pixel level when classifying objects. The main limitation of this study is that proposed uncertainty descriptors can only be calculated based on the labeled data or results of image segmentation prior to prediction.

\section{Method}
\label{sec:Method}

\subsection{The Framework of Interpretable Deep Active Learning for Flood Inundation Mapping}
\label{sec:The Framework of Interpretable Deep Active Learning for Flood Inundation Mapping}

\begin{figure}[t]
  \centering
  \includegraphics[width=1.0 \linewidth]{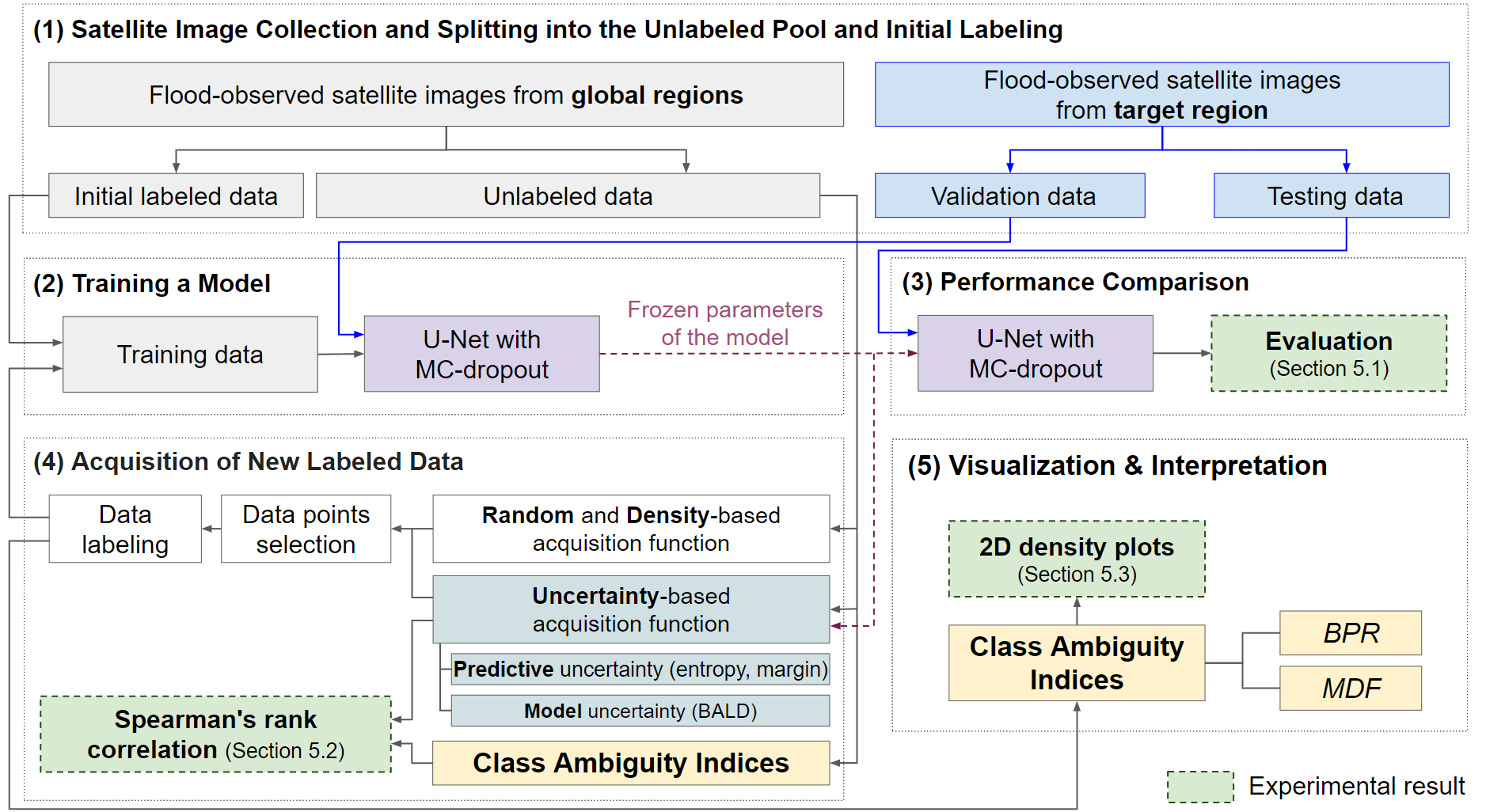}
  \caption{The process of the IDAL-FIM framework.}
  \label{fig:dal-process}
\end{figure}

Deep active learning demonstrates strong predictive performance improvement, but its process is opaque and interpretability is limited due to the black-box nature of deep learning models \citep{goodchild2021replication, hsu2023explainable}. In this study, we introduce the framework of Interpretable Deep Active Learning for Flood Inundation Mapping (IDAL-FIM). The main purpose of the IDAL-FIM framework is to provide interpretation for the deep active learning operation in flood inundation mapping. Fig. \ref{fig:dal-process} illustrates the process of the IDAL-FIM framework. Our proposed framework assumes a pool-based sampling scenario and, accordingly, consists of five stages. In the first stage, satellite images are collected globally to build an unlabeled data pool. The metadata regarding flood events, such as observation period and geographic coordinates, can be obtained from websites of agencies responsible for flood monitoring, including the United Nations Satellite Centre (UNOSAT; https://unosat.org/products/, accessed on 23 April 2024), the Copernicus Emergency Management Service (EMS; https://emergency.copernicus.eu/, accessed on 23 April 2024) and the Dartmouth Flood Observatory (https://floodobservatory.colorado.edu/, accessed on 23 April 2024) \citep{brakenridge2010global}. After that, a small number of satellite images are selected from the unlabeled data pool and annotated to create the initial training data. In addition, for validation and testing purposes, labeled data is generated for satellite images that capture occurrences of flooding in the target area. Then, the satellite images and corresponding labeled data are split into training, validation, and testing data. 

After the initial stage, the following stages are briefly outlined in this section, with a more in-depth explanation presented in separate sections. In the second stage, a deep learning model for flood inundation mapping is trained. In this study, as a deep learning model for segmentation tasks, U-Net with MC-dropout was employed for uncertainty estimation and performance evaluation. More details about the U-Net with MC-dropout are provided in Section 3.3. In the third stage, the performance of the trained deep learning model is evaluated using test data based on the performance metric. Section 4.3 covers detailed configuration for the iteration of deep learning model training and evaluation in the IDAL-FIM framework. In the fourth stage, an acquisition function selects most informative satellite images from an unlabeled data pool. Subsequently, human experts create labeled data using the newly selected satellite images and then they are added to the existing training data. Detailed explanations of the acquisition functions utilized in this study can be found in Section 3.2. Moreover, we demonstrate the statistical significance of the rank correlation between class ambiguity indices and the scores obtained from uncertainty-based acquisition functions in Section 5.2. This statistical analysis aims to support the effectiveness and validity of the class ambiguity indices in interpreting the behavior of active learning. In the final stage, the characteristics of newly labeled multi-spectral satellite images obtained in the prior stage are visualized based on the class ambiguity indices. Two class ambiguity indices, which are Boundary Pixel Ratio (BPR) and Mahalanobis Distance for Flood-segmentation (MDF), are further explained in Section 3.4.

\subsection{Acquisition Functions in the IDAL-FIM Framework}
\label{sec:Acquisition Functions in the IDAL-FIM Framework}

In the IDAL-FIM framework, the acquisition functions take on a pivotal role in the selection of the informative data points, which is associated with reducing the number of labeled data points and improving predictive performance. In this study, we utilized two types of acquisition functions: (1) uncertainty-based and (2) density-based acquisition functions. Specifically, regarding uncertainty-based acquisition function, we leverage the predictive uncertainty and model uncertainty \citep{takezoe2023deep,li2022suggestive,ruuvzivcka2020deep,beluch2018power,gal2017deep,settles2009active}. For predictive uncertainty, entropy \citep{shannon1948mathematical} and margin \citep{scheffer2001active} acquisition functions are employed. On the other hand, for model uncertainty, Bayesian Active Learning by Disagreement \citep[BALD;][]{houlsby2011bayesian} acquisition function is utilized. Furthermore, we implement a density-based acquisition function using Principal Component Analysis (PCA) and K-Means algorithm and include the random acquisition function as a baseline method.

\subsubsection{Uncertainty-based acquisition function}
\label{sec:Uncertainty-based acquisition function }

In the IDAL-FIM framework, a deep learning model for segmentation tasks is trained employing the dropout technique on training data $\mathcal{D}_{train}$. During the inference stage, $T$ forward passes are performed. At each pass, a new dropout mask is sampled, resulting in the model weight $\widehat{\boldsymbol\omega}_t$ at the $t$-th forward pass. In Eq.(\ref{eq:mcd-pred}), the calibrated probability belonging to class $c$, located at ($h$, $w$), is computed as the average of $T$ predicted probabilities \citep{gal2017deep, wang2023calibrating}. The notation $y_{h,w}$ represents the target class at the pixel position ($h$, $w$). $\mathbf{x}$ denotes the input satellite tile image, and $c$ denotes the predefined class.

\begin{align}
\label{eq:mcd-pred}
p(y_{h,w}=c \vert \mathbf{x}, \mathcal{D}_{train}) \approx \frac{1}{T}\sum_{t=1}^T p(y_{h,w}=c \vert \mathbf{x},\widehat{\boldsymbol\omega}_t)
\end{align}

The uncertainty-based acquisition function for segmentation tasks firstly calculates pixel-wise uncertainty by applying each of the uncertainty measures. Then the average uncertainty of all pixels is computed at the tile level. This output value becomes the score of the acquisition function for segmentation tasks and is utilized to determine the priority of data points to be labeled.

\paragraph{Entropy}

One of the most general uncertainty measures is entropy, which is an information-theoretic measure representing the amount of information needed to encode a distribution. Therefore, entropy is commonly perceived as a metric of uncertainty in machine learning \citep{settles2009active,beluch2018power}. In the entropy acquisition function, pixel-level predictive uncertainty is quantified using entropy based on the calibrated class probability in Eq.(\ref{eq:mcd-pred}). Then, the predictive uncertainty located at ($h$, $w$), $u_{h,w}^{Entropy}$, is calculated as in Eq.(\ref{eq:entropy-pixel}):

\begin{align}
\label{eq:entropy-pixel}
u_{h,w}^{Entropy} &= \mathcal{H}[y_{h,w} \vert \mathbf{x},\mathcal{D}_{train}] \nonumber \\ 
                  &= -\sum_{c=1}^C p(y_{h,w}=c \vert \mathbf{x}, \mathcal{D}_{train}) \cdot log\, p(y_{h,w}=c \vert \mathbf{x}, \mathcal{D}_{train}) \nonumber \\ 
                  &\approx -\sum_{c=1}^C(\frac{1}{T}\sum_{t=1}^T p(y_{h,w}=c \vert \mathbf{x},\widehat{\boldsymbol\omega}_t)) \cdot log(\frac{1}{T}\sum_{t=1}^T p(y_{h,w}=c \vert \mathbf{x},\widehat{\boldsymbol\omega}_t))
\end{align}
where $\mathcal{H}$ denotes entropy \citep{shannon1948mathematical}. The score of the entropy acquisition function for  segmentation $s^{Entropy}$ is the average of the pixel-level predictive uncertainties, and as the score becomes higher, the priority of selecting the data points also increases.

\begin{align}
\label{eq:entropy-tile}
s^{Entropy} &= \frac{1}{HW} \sum_{h=1}^H \sum_{w=1}^W u_{h,w}^{Entropy}
\end{align}

\paragraph{Margin}
The margin is defined as the difference between the probabilities of the two most probable classes. A higher margin indicates that the model’s prediction is more certain, with lower predictive uncertainty, as the probabilities of the two most probable classes are more distinct \citep{scheffer2001active,beluch2018power}. Notably, in the binary classification setting, the margin function reduces to the entropy function, as both are equivalent to querying the instance with a class posterior closest to 0.5 \citep{settles2009active}. Therefore, the characteristics of the margin acquisition function and the entropy acquisition function become analogous in binary segmentation problems. The pixel-level uncertainty measured by margin located at ($h$, $w$), $u_{h,w}^{Margin}$, is calculated as in Eq.(\ref{eq:margin-pixel}) where c1 and c2 are the first and second most probable class labels under the model, respectively.
\begin{align}
\label{eq:margin-pixel}
u_{h,w}^{Margin} = \frac{1}{T}\sum_{t=1}^Tp(y_{h,w}=c_1 \vert \mathbf{x},\widehat{\boldsymbol\omega}_t) - \frac{1}{T}\sum_{t=1}^Tp(y_{h,w}=c_2 \vert \mathbf{x},\widehat{\boldsymbol\omega}_t) 
\end{align}
The score of the margin acquisition function for segmentation $s^{Margin}$ is calculated according to Eq.(\ref{eq:margin-tile}). As the score of the margin acquisition function decreases, those data points are prioritized for selection.

\begin{align}
\label{eq:margin-tile}
s^{Margin} &= \frac{1}{HW} \sum_{h=1}^H \sum_{w=1}^W u_{h,w}^{Margin}
\end{align}

\paragraph{BALD (Bayesian Active Learning by Disagreement)}
BALD is defined as the mutual information between predictions and model posterior. This means that the value of BALD-based function is maximized when the model generates uncertain predictions on average and also confidently produces disagreeing predictions simultaneously \citep{gal2017deep}. Consequently, BALD was utilized as the measure of model uncertainty in the previous study \citep{jesson2021causal} because it highlights the variability in class probabilities across different stochastic forward passes \citep{gal2017deep}. The pixel-level uncertainty using BALD located at ($h$, $w$), $u_{h,w}^{BALD}$, and the score of the BALD acquisition function for segmentation, $s^{BALD}$, are calculated as in Eq.(\ref{eq:bald-pixel}) and (\ref{eq:bald-tile}). 

\begin{align}
\label{eq:bald-pixel}
   u_{h,w}^{BALD} &= \mathcal{H}[y_{h,w} \vert \mathbf{x},\mathcal{D}_{train}] - \mathbb{E}_{p(\boldsymbol\omega \vert \mathcal{D}_{train})}[\mathcal{H}[y_{h,w} \vert \mathbf{x}, \boldsymbol\omega]] \nonumber \\
                  &\approx -\sum_{c=1}^C(\frac{1}{T}\sum_{t=1}^T p(y_{h,w}=c \vert \mathbf{x},\widehat{\boldsymbol\omega}_t)) \cdot log(\frac{1}{T}\sum_{t=1}^T p(y_{h,w}=c \vert \mathbf{x},\widehat{\boldsymbol\omega}_t)) \nonumber \\
                  &\quad -\frac{1}{T}\sum_{t=1}^T \sum_{c=1}^C -p(y_{h,w}=c \vert \mathbf{x},\widehat{\boldsymbol\omega}_t) \cdot log\,p(y_{h,w} =c \vert \mathbf{x},\widehat{\boldsymbol\omega}_t) 
\end{align}

\begin{align}
\label{eq:bald-tile}
s^{BALD} &= \frac{1}{HW} \sum_{h=1}^H \sum_{w=1}^W u_{h,w}^{BALD}
\end{align}

\subsubsection{Density-based acquisition function}
\label{sec:Density-based acquisition function}

In order to compare characteristics with uncertainty-based acquisition functions in the framework of IDAL-FIM, we implemented a simple density-based acquisition function using Principal Component Analysis (PCA) and K-Means under the assumption that data points maximizing data feature diversity are informative for training deep learning models \citep{xie2020deal,takezoe2023deep}. The density-based acquisition function used in this study has two steps. First, PCA was performed to reduce the dimensionality of the unlabeled multi-spectral satellite images. Here, the output of PCA is considered to be the features of the unlabeled data. Then, using the PCA output as input, the K-means algorithm identified $k$ new samples closest to the centroid of each cluster. The rationale behind selecting new samples closest to the centroids formed by the K-means algorithm for each cluster is to maximize the diversity of features. This is because the objective of the K-means algorithm is to find clusters that are internally coherent but maximally distinct from each other. We therefore name this acquisition function as K-means acquisition function. 

\SetKwComment{Comment}{/* }{ */}
\DontPrintSemicolon
\RestyleAlgo{ruled}
\begin{algorithm}
\caption{Pseudocode of density-based acquisition function utilizing PCA and K-Means}\label{alg:two}
\KwData{ $NewSamples$, $Data_{ReducedDim}$, $Data_{UnlabeledSat}$, $N_{ReducedDim}$, $Clusters$ }
\KwResult{$NewSamples$}
$NewSamples = [ \, ]$\;
$Data_{ReducedDim} =$ Perform PCA on $Data_{UnlabeledSat}$ to reduce its dimension to $N_{ReducedDim}$\;
$Clusters =$ Apply the K-means algorithm to cluster $Data_{ReducedDim}$\;
\For{$cluster \in Clusters$}{
  Calculate the centroid of the $cluster$ \;
  Find the $sample$ closest to the centroid within the $cluster$ \;
  $NewSamples \longleftarrow NewSamples \cup sample$\;
}
\end{algorithm}

\subsection{Deep Learning Model for Segmentation Task in the IDAL-FIM Framework}
\label{sec:Deep Learning Model for Segmentation Task in the IDAL-FIM Framework}

The efficiency and effectiveness of the deep active learning framework are closely linked to the choice of both a deep learning model and an uncertainty estimation method due to the iterative nature of active learning. We selected the U-Net as the deep learning model for segmentation within the IDAL-FIM framework, which was utilized in a recent uncertainty estimation study of water body mapping \citep{hertel2023probabilistic}. Furthermore, we considered the following three factors to determine the uncertainty estimation method suitable for the IDAL-FIM framework. First, the computational cost of uncertainty estimation should be low since training and inference of the deep learning model are repeatedly conducted within the IDAL-FIM framework. MC-dropout achieves a lower computation cost by training a single deep learning model and performing multiple inferences, compared to deep ensembles and fully Bayesian CNNs. Second, as the uncertainty-based acquisition function for semantic segmentation computes the average uncertainty across all pixels, the higher reliability of uncertainty estimation at a pixel-level holds relatively less importance in the IDAL-FIM framework. Lastly, the ease of incorporating other deep learning models into the proposed framework was also considered.

\begin{figure}[t]
  \centering
  \includegraphics[width=1.0 \linewidth]{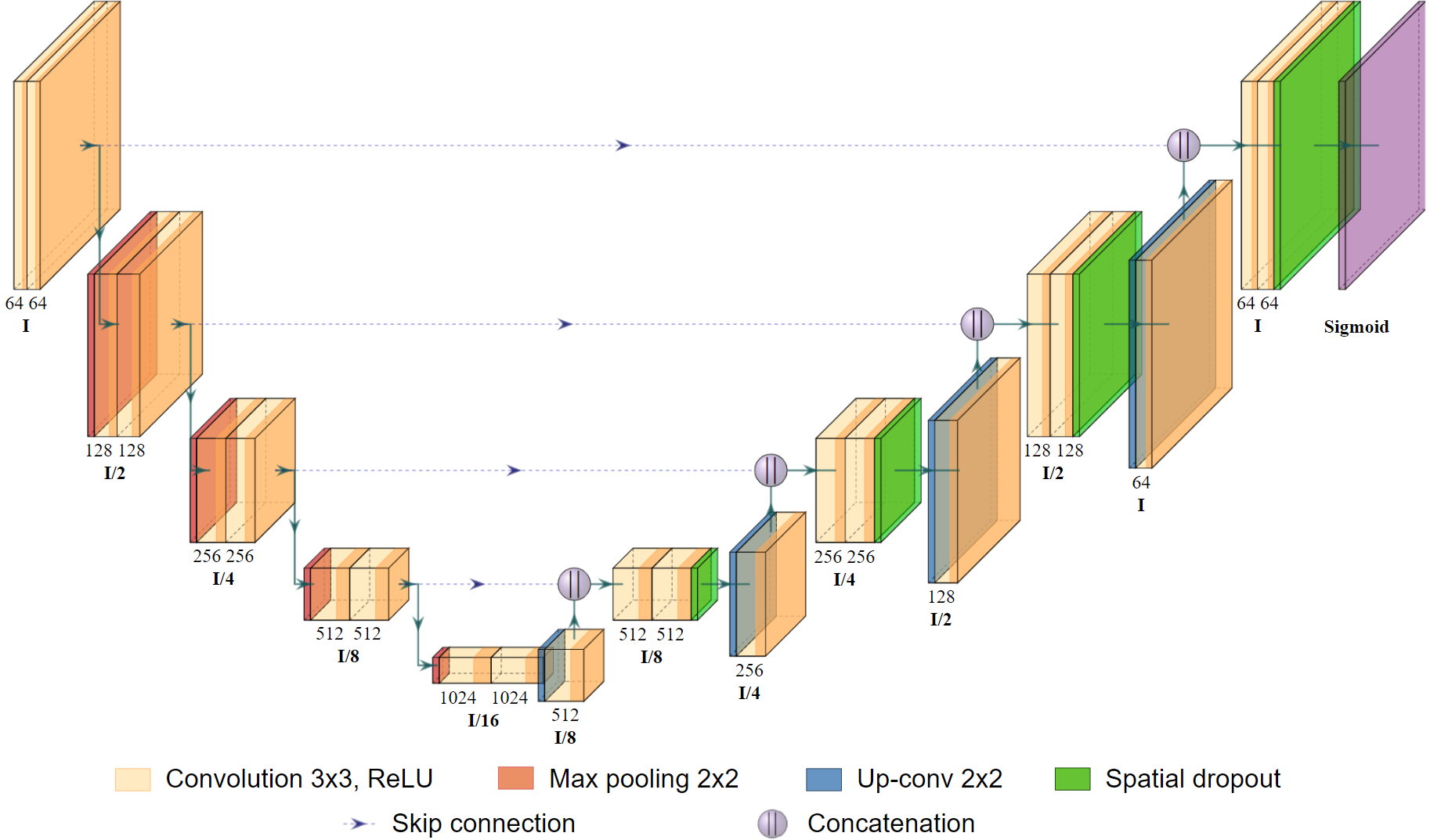}
  \caption{The architecture of U-Net with MC-dropout. The input image assumes uniform width and height (I).}
  \label{fig:arch-bayes-unet}
\end{figure}

Fig. 2 illustrates the architecture of the U-Net with MC-dropout utilized in this study. This model consists of two main components: the encoder and the decoder. Same as the existing U-Net model, the encoder utilizes a down-sampling process four times to extract features and reduce computational cost. This down-sampling process is composed of two convolutional layers that increase the number of channels and a pooling layer that decreases the spatial resolution. These extracted features are then forwarded to the decoder, which has a symmetric structure to the encoder and employs a four times up-sampling process to reconstruct the spatial information of the input. U-Net with MC-dropout also integrates skip connections to capture precise locations at each step of the decoder. These skip connections include concatenating the output of the decoder layers with the corresponding feature maps from the encoder at the same level, thereby enhancing the precision of pixel segmentation. Regarding MC-dropout, instead of using regular dropout, spatial dropout is applied at the end of the up-sampling process. Spatial dropout is a regularization technique in deep learning where specific proportions of two-dimensional feature maps are randomly set to zero on a per-channel basis during training to enhance model robustness and prevent over-fitting \citep{tompson2015efficient}.

\subsection{Class Ambiguity Indices in the IDAL-FIM Framework }
\label{sec:Class Ambiguity Indices in the IDAL-FIM Framework}

We present two class ambiguity indices to quantify the tile-level binary inter-class ambiguity of the input satellite image. For calculation of class ambiguity indices, pixel-wise labeled data is required, similar to the uncertainty descriptors in previous study \citep{zhang2019uncertainty}. However, our proposed indices are utilized after the labeling stage, therefore, those limitations are not relevant to this study.  

The two class ambiguity indices are designed to quantify the class ambiguity between flood and non-flood class stemming from the spatial and spectral resolution constraints of the sensor in input satellite images. The proposed two class ambiguity indices are (1) Boundary Pixel Ratio (BPR) and (2) Mahalanobis Distance for Flood-segmentation (MDF). BPR is designed to represent ambiguity between flooded and non-flooded classes due to spatial resolution constraints and is calculated as the proportion of boundary pixels within satellite images:
\begin{align}
    \text{BPR} = \frac{\text{BPS}}{\text{TPS}} 
\end{align}
where BPS is the total number of boundary pixels, and TPS is the total number of pixels in a satellite image. Boundary pixel is defined as the class of the center pixel that differs from at least one of its surrounding eight pixels. Therefore, increasing BPR implies more pixels with higher inter-class ambiguity due to the spatial resolution limitations inherent in satellite imagery.

On the other hand, MDF is formulated to capture the semantic ambiguity between flooded and non-flooded classes at the tile level. This is calculated as the Mahalanobis distance of average pixel values between flood and non-flood class. Mahalanobis distance \citep{mahalanobis1936genaralised} is a measure of the distance between points over a given distribution. Therefore, we assumed that the pixel values of each class follow a multivariate normal distribution. Additionally, decreasing MDF suggests increased semantic ambiguity between flooded and non-flooded areas at the tile level, attributed to the uncertainty in terms of spectral similarity.
\begin{align}
    \text{MDF} = \sqrt{(\mathbf{p}_{\text{flood}} - \mathbf{p}_{\text{non-flood}})^T\Sigma^{-1}(\mathbf{p}_{\text{flood}} - \mathbf{p}_{\text{non-flood}})}
\end{align}
where $\mathbf{p}_{\text{flood}}$ and $\mathbf{p}_{\text{non-flood}}$ is a vector of average pixel values in each class, which are flood and non-flood, and $\Sigma$ is positive-definite covariance matrix with rows and columns matching the number of channels in the input satellite images. In addition to the two class ambiguity indices, we employ the Flood Pixel Ratio (FPR) as a class imbalance index to interpret the capability of mitigating class imbalance issues in the IDAL-FIM framework:
\begin{align} 
    \text{FPR} = \frac{\text{FPS}}{\text{TPS}} 
\end{align} 
where FPS is the number of flood pixels.   

\section{Experimental Setup}
\label{sec:Experimental Setup}

\subsection{Dataset and Data Preprocessing}
\label{sec:Dataset and Data Preprocessing}

Sen1Floods11, a georeferenced flood inundation mapping dataset \citep{bonafilia2020sen1floods11}, was utilized in this experiment. This dataset includes 446 pairs of image data, consisting of satellite imagery from Sentinel-1 and Sentinel-2, along with corresponding labeled data generated by experts for flood inundation mapping. Satellite imagery from Sentinel-1 and Sentinel-2 in Sen1Floods11 captures 11 flood events occurring across various countries worldwide between 2016 and 2019. Each of the satellite images has a 10 meter resolution and dimensions of 512 $\times$ 512 pixels. The satellite imagery observed by Sentinel-1 consists of SAR (Synthetic Aperture Radar) imagery, which is composed of VH and VV bands. The imagery captured by Sentinel-2 comprises multi-spectral imagery with 13 bands, including red, green, blue, Near InfraRed (NIR), and Short-Wave Infrared (SWIR). All bands of the multi-spectral images are linearly interpolated to 10 meters to ensure uniform spatial resolution. 

\begin{figure}[t]
  \centering
  \includegraphics[width=0.8 \linewidth]{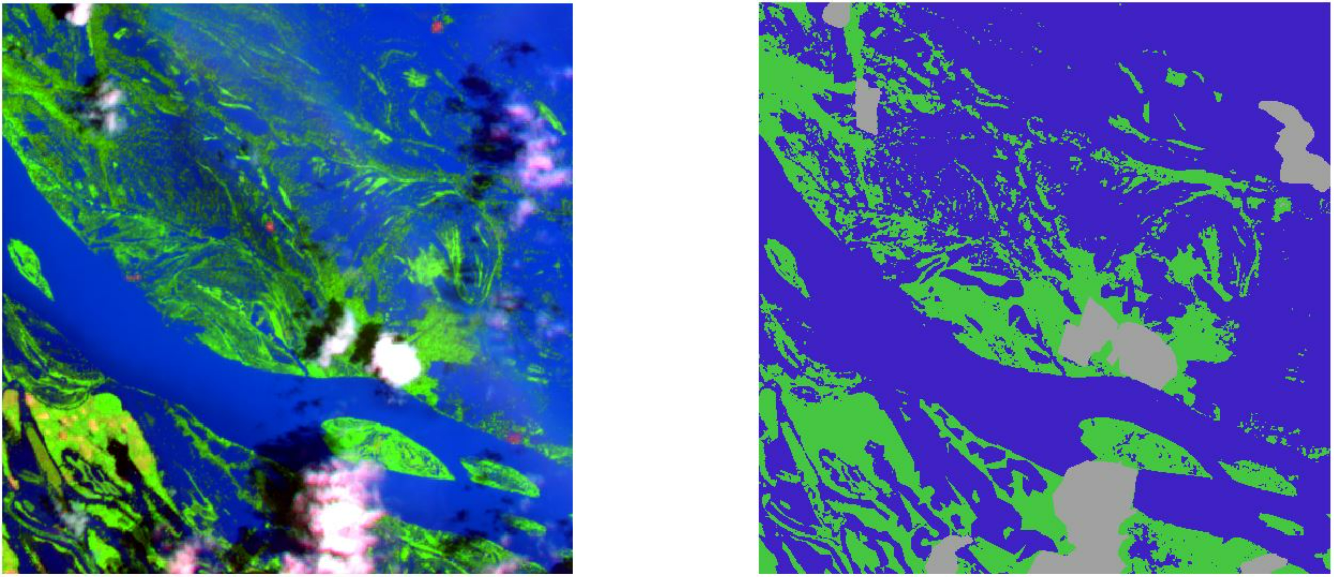}
  \caption{Example images in Sen1Floods11. (left) Sentinel-2 false color composite image, and (right) corresponding labeled data with color codes: blue for flood, green for non-flood, and gray for no data.}
  \label{fig:sample-sen1floods11}
\end{figure}

Sen1Floods11 has been utilized in several studies focusing on deep learning models for flood inundation mapping \citep{konapala2021exploring,bai2021enhancement,katiyar2021near,yadav2022attentive}. In particular, we drew upon the findings of \citet{konapala2021exploring} for input bands selection and data preprocessing. The authors investigated the optimal input band combination in U-Net for flood inundation mapping based on Sen1Floods11. The authors showed that utilizing multi-spectral satellite imagery as input led to a higher F1-score compared to SAR imagery. This result suggests that multi-spectral imagery provides more advantages over SAR imagery in automating the process and diminishing the necessity for expert corrections in flood inundation mapping, especially in cases with minimal cloud cover. As a result, we employ multi-spectral satellite imagery in Sen1Floods11 as input for this study. 

Regarding data preprocessing, \citet{konapala2021exploring} reported that HSV (Hue, Saturation, Value) conversion using the red, NIR, and SWIR2 bands is effective for flood inundation mapping through their experiment. Hence, HSV transformed values based on red, NIR, and SWIR2 were employed as a data preprocessing for our experiments. In addition, the 512 $\times$ 512 multi-spectral satellite images in Sen1Floods11 were divided into four non-overlapping 256 $\times$ 256 pixel tiles for the efficient GPU memory utilization. 

\subsection{Data Splitting}
\label{sec:Data Splitting}

Among the 11 regions in Sen1Floods11, multi-spectral satellite images from 8 regions (Ghana, India, Pakistan, Paraguay, Somalia, Spain, Sri-Lanka, and USA) were used for the pool of unlabeled data, and the remaining 3 regions (Bolivia, Nigeria, and Vietnam) were designated for the target region. This split was taken into consideration of their geographic locations, as depicted in Fig. 4. The experiment is designed to carry out flood inundation mapping within the IDAL-FIM framework, targeting the regions of Bolivia, Nigeria, and Vietnam, and using the same unlabeled data pool. In each target region, the multi-spectral satellite images and corresponding labeled data are randomly split, with 50\% allocated for validation and 50\% for testing. Initial labeled data were randomly selected using a fixed seed number in a single experiment.

\begin{figure}[t]
  \centering
  \includegraphics[width=1.0 \linewidth]{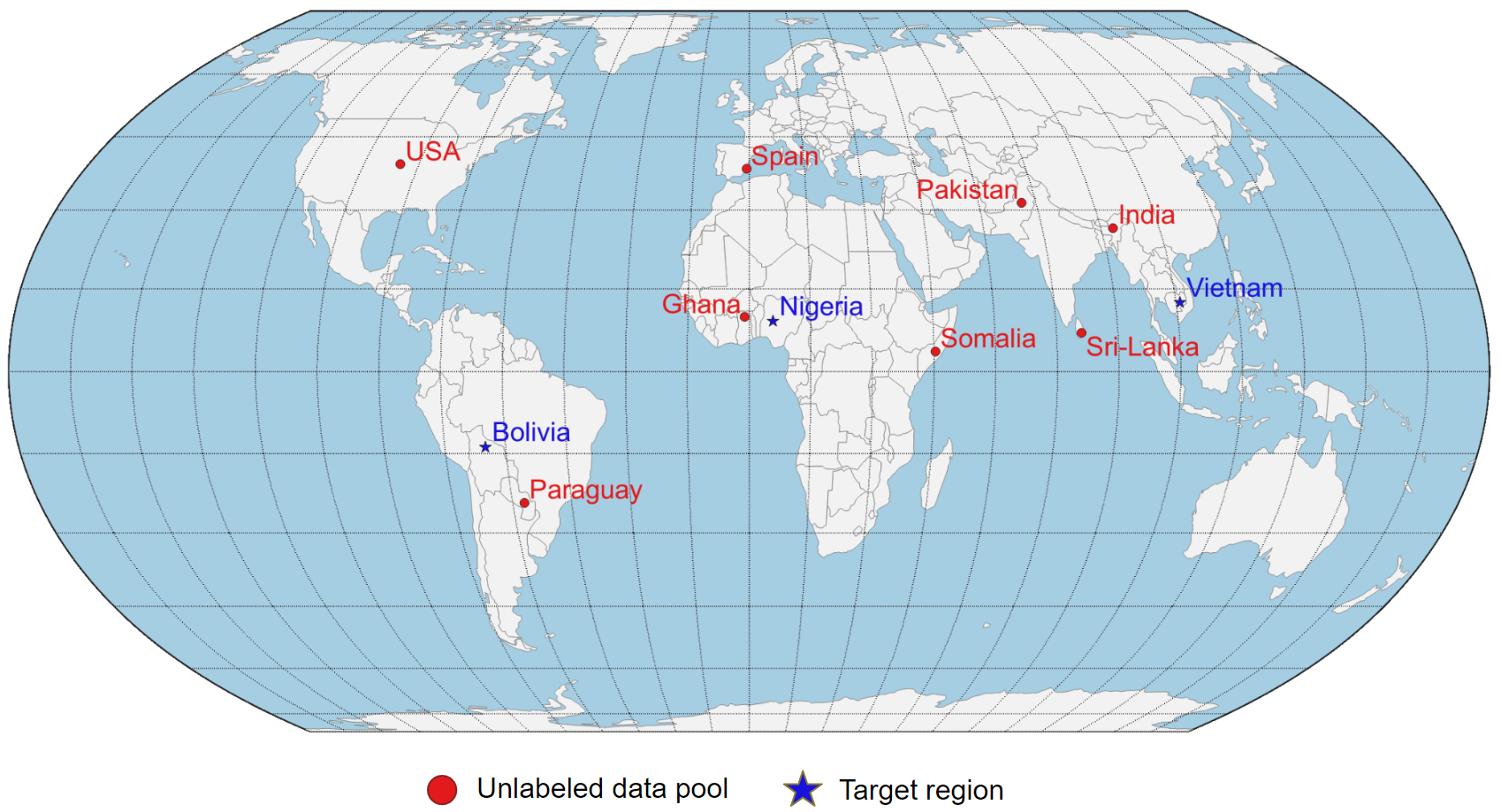}
  \caption{The geographical regions that make up the unlabeled data pool and the target regions. The target regions were selected, one from each of the continents of South America, Africa, and Asia. The remaining 8 regions were utilized for the unlabeled data pool.}
  \label{fig:data-geo-dist}
\end{figure}

\begin{table}[h]
\centering
\caption{The number of multi-spectral satellite images in the unlabeled data pool}
\label{tab:The number of multispectral satellite images in the unlabeled data pool}
\resizebox{\textwidth}{!}{%
\begin{tabular}{@{}cccccccccc@{}}
\toprule
Region & Total & Ghana & India & Pakistan & Paraguay & Somalia & Spain & Sri-Lanka & USA \\ \midrule
Count  & 1,532 & 212   & 272   & 112      & 268      & 104     & 120   & 168       & 276 \\ \bottomrule
\end{tabular}%
}
\end{table}

\begin{table}[h]
\centering
\caption{The number of multi-spectral satellite images in the target regions}
\label{tab:The number of multispectral satellite images in the target regions}
\resizebox{0.48\textwidth}{!}{%
\begin{tabular}{@{}cclclclc@{}}
\toprule
Region & \multicolumn{2}{c}{Total} & \multicolumn{2}{c}{Bolivia} & \multicolumn{2}{c}{Nigeria} & Vietnam \\ \midrule
Count  & \multicolumn{2}{c}{252}   & \multicolumn{2}{c}{60}      & \multicolumn{2}{c}{72}      & 120     \\ \bottomrule
\end{tabular}%
}
\end{table}

\subsection{The Configurations for Iterative Training and Evaluation}
\label{sec:The Configurations for Iterative Training and Evaluation}

This section explains the detailed configurations for iterative training and evaluation within the IDAL-FIM framework. Table \ref{tab:Components and parameters of DAL-FIM framework} shows the components and parameters of the IDAL-FIM framework for experiments. During experiments, labeled data in Sen1Floods11 are considered to be created by human experts in the data labeling stage of the IDAL-FIM framework. For reproducibility, the settings related to randomness (e.g. weight initialization, dataset shuffling, nondeterministic algorithms, etc.) were configured to guarantee consistent outputs using the same random seed. The number of initial training data, newly acquired samples per iteration, and number of iterations were determined considering the size of the unlabeled data pool used in the experiment. 

Furthermore, during each iteration, the U-Net with MC-dropout is trained from scratch using the hyperparameters specified in Table \ref{tab:Hyperparameters for training Bayesian U-Net}. Random flip was applied for data augmentation. Early stopping is employed when the validation loss does not improve for the specified number of epochs. For carrying out the experiment, PyTorch 1.8.1 and the following hardware was used for all processing: Intel Xeon E5 1.9GHz (72 Cores), 64GB 2666MHz DDR4, 1.5 TB HD, Nvidia GeForce GTX 980 Ti 24GB.

\begin{table}[h]
\centering
\caption{Components and parameters of IDAL-FIM framework}
\label{tab:Components and parameters of DAL-FIM framework}
\resizebox{\textwidth}{!}{%
\begin{tabular}{@{}lll@{}}
\toprule
\multicolumn{2}{l}{\textbf{Components / Parameters}}                                            & \textbf{Value} \\ \midrule
Component &
  \begin{tabular}[c]{@{}l@{}}Acquisition Functions \\ (Section \ref{sec:Acquisition Functions in the IDAL-FIM Framework})\end{tabular} &
  \begin{tabular}[c]{@{}l@{}}Baseline (Random), Uncertainty-based\\ (Entropy, Margin, BALD), Density-based (K-Means)\end{tabular} \\ \cmidrule(l){2-3} 
 &
  \begin{tabular}[c]{@{}l@{}}Deep learning model \&\\ uncertainty estimation method\\ (Section \ref{sec:Deep Learning Model for Segmentation Task in the IDAL-FIM Framework})\end{tabular} &
  U-Net / MC-dropout \\ \cmidrule(l){2-3} 
 &
  \begin{tabular}[c]{@{}l@{}}Class ambiguity and imbalance\\Indices (Section \ref{sec:Class Ambiguity Indices in the IDAL-FIM Framework})\end{tabular} &
  \begin{tabular}[c]{@{}l@{}}BPR, MDF, FPR\end{tabular} \\ \cmidrule(l){2-3} 
          & \begin{tabular}[c]{@{}l@{}}Dataset\\ (Section \ref{sec:Dataset and Data Preprocessing})\end{tabular}                     & \begin{tabular}[c]{@{}l@{}}Sen1Floods11 (Multi-spectral satellite imagery and its \\ corresponding labeled data)\end{tabular} \\ \midrule
Parameter & \begin{tabular}[c]{@{}l@{}}Initial training data\end{tabular} & 100 (random selection)            \\ \cmidrule(l){2-3} 
          & Initial unlabeled data pool                                                         & 1,532          \\ \cmidrule(l){2-3} 
 &
  Validation and testing data &
  \begin{tabular}[c]{@{}l@{}}50\% of the data in the target region is allocated \\ for validation, and the other 50\% is allocated for testing\end{tabular} \\ \cmidrule(l){2-3} 
          & \begin{tabular}[c]{@{}l@{}}Newly acquired samples \\ per iteration\end{tabular}        & 100            \\ \cmidrule(l){2-3} 
          & Number of iterations                                                             & 4              \\ \cmidrule(l){2-3} 
          & \begin{tabular}[c]{@{}l@{}}Number of total runs\end{tabular}        & 10             \\ \bottomrule
\end{tabular}%
}
\end{table}

\begin{table}[h]
\centering
\caption{Hyperparameters for training U-Net with MC-dropout}
\label{tab:Hyperparameters for training Bayesian U-Net}
\resizebox{0.75\textwidth}{!}{%
\begin{tabular}{@{}ll@{}}
\toprule
\textbf{Hyperparameter}                    & \textbf{Value}       \\ \midrule
Loss function                               & Binary Cross Entropy \\
Optimizer                                   & AdamW                \\
Learning rate                               & 5e-4                 \\
Weight decay                                & 1e-2                 \\
Maximum epoch                               & 300                  \\
Batch size                                  & 8                    \\
The number of inferences through MC-dropout & 10                   \\
Spatial dropout rate                        & 0.5                  \\ 
\begin{tabular}[c]{@{}l@{}}Early stopping \\ (monitoring variable / delta / patience)\end{tabular} & Validation loss / 5e-4 / 5 \\ \bottomrule
\end{tabular}%
}
\end{table}

To evaluate the performance in each iteration of the IDAL-FIM framework, the F1-score is reported using a prediction probability threshold of 0.5 \citep{wieland2023semantic}. Cross-validation is not taken into consideration due to the computational overhead involved in performing each iteration multiple times. The F1-score is calculated using True Positive (TP), False Positive (FP), False Negative (FN) in a confusion matrix. When calculating the confusion matrix for the flood class, the no-data class in the input multi-spectral satellite image was considered as a non-flood class. 

\begin{align}
    \text{Precision} = \frac{\text{TP}}{\text{TP} + \text{FP}} 
\end{align}
\begin{align}
    \text{Recall} = \frac{\text{TP}}{\text{TP} + \text{FN}}
\end{align}
\begin{align}
    \text{F1-score} = \frac{2 \times \text{Precision} \times \text{Recall}}{\text{Precision}  + \text{Recall}} 
\end{align}

In addition, as the same experiment is repeated multiple times for each acquisition function, the mean F1-score (mF1-score) and the standard deviation of F1-score (sdF1-score) is calculated as follows:

\begin{align}
    \text{mF1-score} = \frac{1}{N}\sum_{i=1}^N \text{F1-score}_i
\end{align}
\begin{align}
    \text{sdF1-score} = \sqrt{\frac{1}{N}\sum_{i=1}^N (\text{F1-score}_i - \text{mF1-score})^2}
\end{align}

\section{Result}
\label{sec:Result}

\subsection{Evaluation on Model Performance with Varying Acquisition Functions and Training Data Sizes}
\label{sec:Evaluation on Model Performance with Varying Acquisition Functions and Training Data Sizes}

First, we conducted experiments to evaluate the impact of active learning strategies on model performance. The F1-score of the model was measured in the evaluation stage during the iterative process of the IDAL-FIM framework. For the convenience of notation regarding the model, Model$_{\text{AF-N}}$ denotes the model trained on N data points selected by the acquisition function AF, and Model$_{\text{Full}}$ refers to the model trained on the entire 1,532 data points in the unlabeled data pool. We compared five acquisition functions for their effectiveness in terms of the mean F1-score across ten experiments. As depicted in Fig. 5, we utilized one baseline and one upper bound mean F1-scores: mF1-score$_{\text{Random-500}}$, which is the mean F1-score of Model$_{\text{Random-500}}$, displayed as a horizontal black dashed line as a baseline, and mF1-score$_{\text{Full}}$, which is the mean F1-score of Model$_{\text{Full}}$, represented as a horizontal blue dashed line as a upper bound performance. 

Fig. 5 shows that the mean F1-score of the models, which are trained on the data points acquired based on the predictive uncertainty such as margin and entropy, consistently achieved the most comparable mean F1-scores to the mF1-score$_{\text{Full}}$. This result shows that a model trained on a subset of the entire dataset selected by the margin and entropy acquisition function can achieve equivalent performance to a model trained on the entire dataset. Additionally, the mean F1-score of models trained on the data points selected by the margin acquisition function outperformed that of the random acquisition function across the three regions. In Bolivia, both the margin and BALD acquisition function achieved a superior mean F1-score despite having 300 fewer training data points. Similarly, in Nigeria, the margin and entropy acquisition function outperformed the random acquisition function despite having 300 fewer training data points. In Vietnam, the margin acquisition function surpassed despite having 200 fewer training data points. In each experiment, we used the same random seed to ensure the performance evaluation in an identical environment. Therefore, the mean F1-score of all five acquisition functions has the same value when the number of training data is 100 in each of the three regions.

\begin{figure}[t]
  \centering
  \includegraphics[width=1.0 \linewidth]{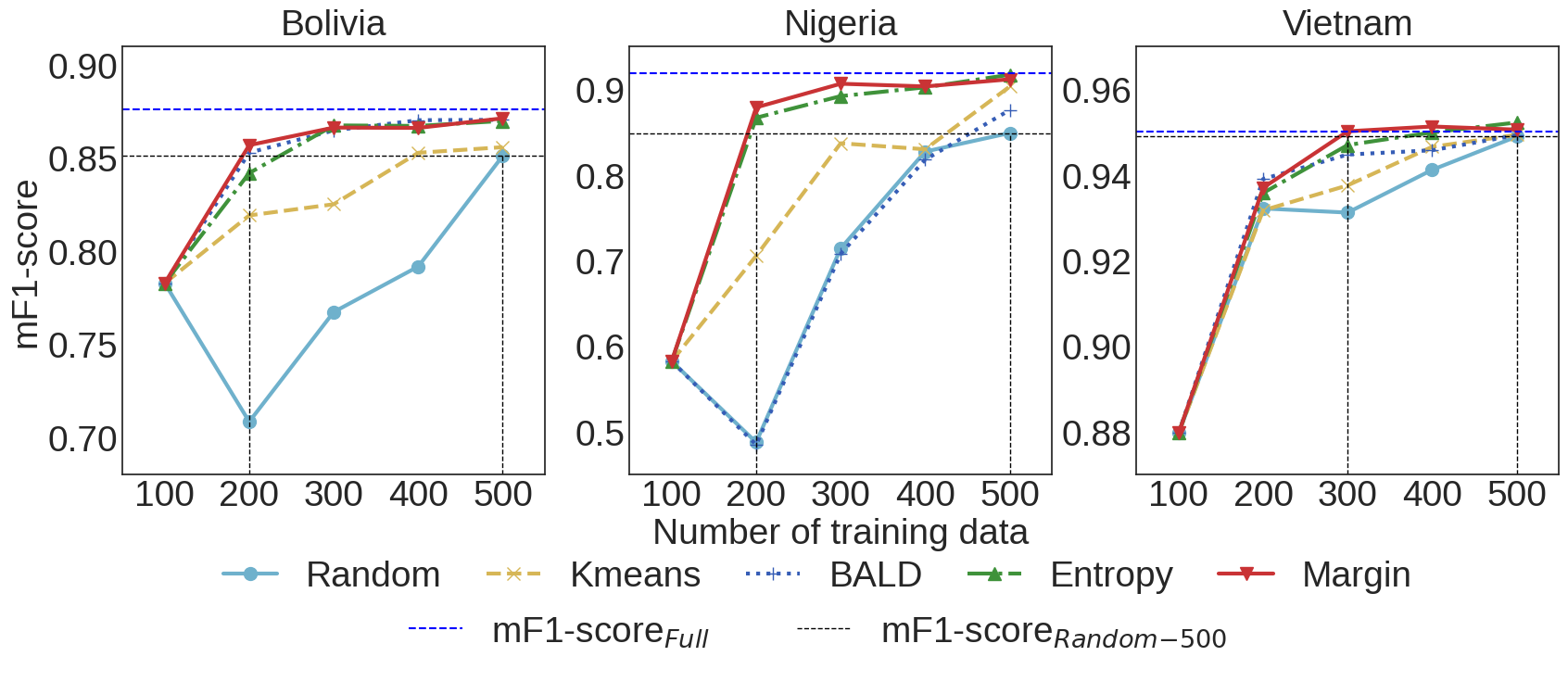}
  \caption{The comparison of the mean F1-score in different five acquisition functions: (left) Bolivia, (middle) Nigeria, (right) Vietnam. The horizontal blue dashed line (mF1-score$_{\text{Full}}$) represents the mean F1-score from models trained on the entire 1,532 data points in the pool. The horizontal black dashed line (mF1-score$_{\text{Random-500}}$) displays the mean F1-score from models trained on a selection of 500 data points using the random acquisition function.}
  \label{fig:mean-f1-score}
\end{figure}

In this experiment, we expected that increasing the amount of training data would lead to a gradual improvement in mean F1-scores. However, we observed a degradation in mean F1-scores at specific points in Fig. 5 when utilizing the random acquisition function. Degradation was particularly observed with training data points of 200 in Bolivia and Nigeria, and with training data points of 300 in Vietnam. To investigate the cause of the decrease in the mean F1-score at a specific iteration, we examined the standard deviation of the F1-score for each iteration. As shown in Fig. 6, we observed a tendency for the mean F1-score to not consistently increase when the standard deviation of the F1-score becomes relatively higher compared to other acquisition functions as each iteration progresses. Particularly, in the case of the random acquisition function, the standard deviation of the F1-score was larger than the other four acquisition functions when the number of training data ranged from 200 to 400. On the contrary, in the case of the margin acquisition function, which was the best-performing acquisition function within the IDAL-FIM framework, it tended to show a more rapid decrease in the standard deviation of the F1-score as the number of training data points increases, compared to other acquisition functions. 

\begin{figure}[t]
  \centering
  \includegraphics[width=1.0 \linewidth]{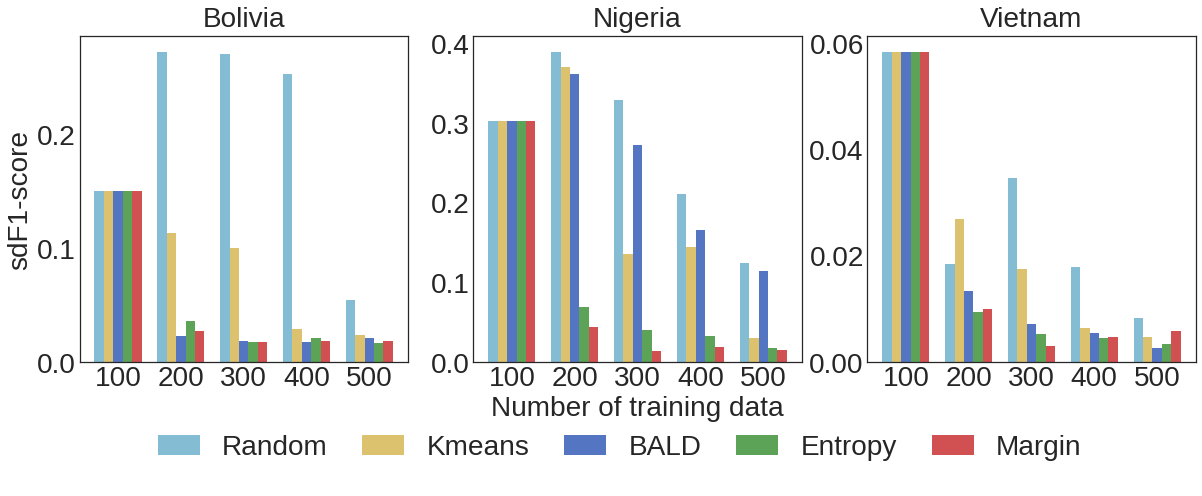}
  \caption{The comparison on the standard deviation of F1-score in different five acquisition functions: (left) Bolivia, (middle) Nigeria, (right) Vietnam.}
  \label{fig:sd-f1-score}
\end{figure}

Following the overall performance assessment of acquisition functions within the IDAL-FIM framework, in Fig. 7, we examined visual examples of prediction results from each target region. In this visualization, results from three models are compared: (1) Model$_{\text{Random-500}}$, (2) Model$_{\text{Margin-500,}}$ and (3) Model$_{\text{Full}}$. Model$_{\text{Margin-500,}}$ is the model train on the data points selected by best-performed acquisition function in the IDAL-FIM framework, and Model$_{\text{Random-500}}$ is a baseline model. Model$_{\text{Full}}$ is a model representing upper bound performance. In Fig. 7, prediction results of Model$_{\text{Margin-500,}}$ and Model$_{\text{Full}}$ are consistently similar across three regions (a), (b), and (c) in terms of True Positive (TP), True Negative (TN), False Positive (FP) and False Negative (FN). On the other hand, the prediction result of Model$_{\text{Random-500}}$ exhibited more false positive pixels in Bolivia (Fig. 7 (a)) and Nigeria (Fig. 7 (b)) compared to Model$_{\text{Margin-500}}$ and Model$_{\text{Full}}$.

\begin{figure}[t]
  \centering
  \includegraphics[width=1.0 \linewidth]{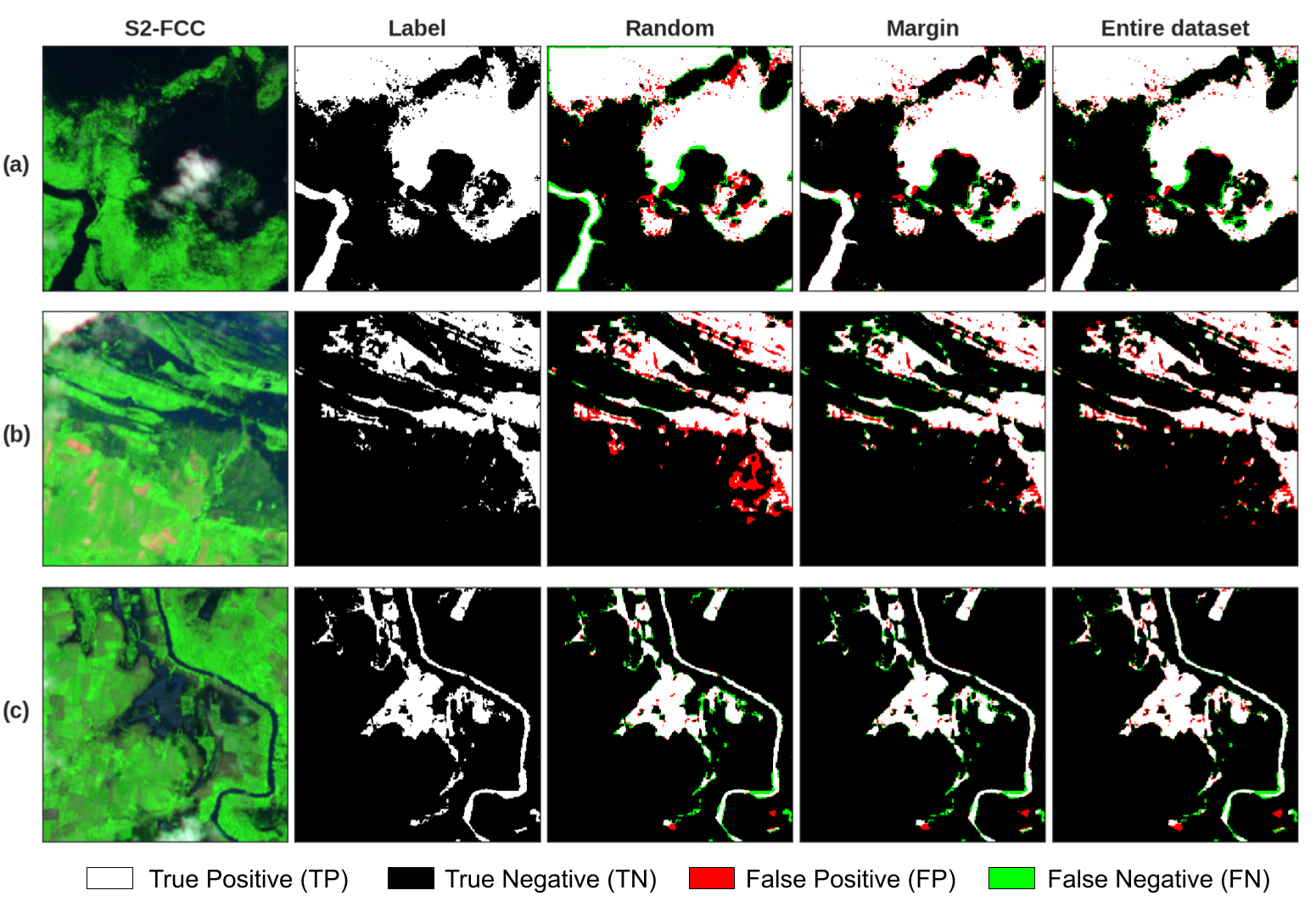}
  \caption{The comparison on prediction results of examples in each region: (a) Bolivia, (b) Nigeria, (c) Vietnam. ``S2-FCC'' is the false color composite of input sentinel-2 image using red, NIR and SWIR. ``Label'' has flood (white) and non-flood (black) pixel information. ``Random'' and ``Margin'' are the prediction results from a Model$_{\text{Random-500}}$ and Model$_{\text{Margin-500}}$, respectively. ``Entire dataset'' represents the prediction results from a Model$_{\text{Full}}$. 
}
  \label{fig:pred-sample}
\end{figure}

\subsection{Relationship Between Class Ambiguity Indices and the Score of Uncertainty-based Acquisition Functions}
\label{sec:Relationship Between Class Ambiguity Indices and the Score of Uncertainty-based Acquisition Functions}

We investigated the relationship between class ambiguity indices and the Uncertainty-based Acquisition Function (UAF) score, which is directly associated with the average uncertainty of all pixels, by calculating the correlation coefficient at each iteration within the IDAL-FIM framework. In each iteration, at the stage of acquiring new labeled data, we calculate the two class ambiguity indices, which are the BPR and the MDF, as well as scores of the UAF, for all the data points in the unlabeled pool. Then, we obtain the Spearman’s rank correlation coefficient between the BPR and the UAF score, and between the MDF and the UAF score. The reason we chose the Spearman’s rank correlation coefficient is because the UAF scores determine the ranking of selected data points.

\begin{figure}[t]
  \centering
  \includegraphics[width=1.0 \linewidth]{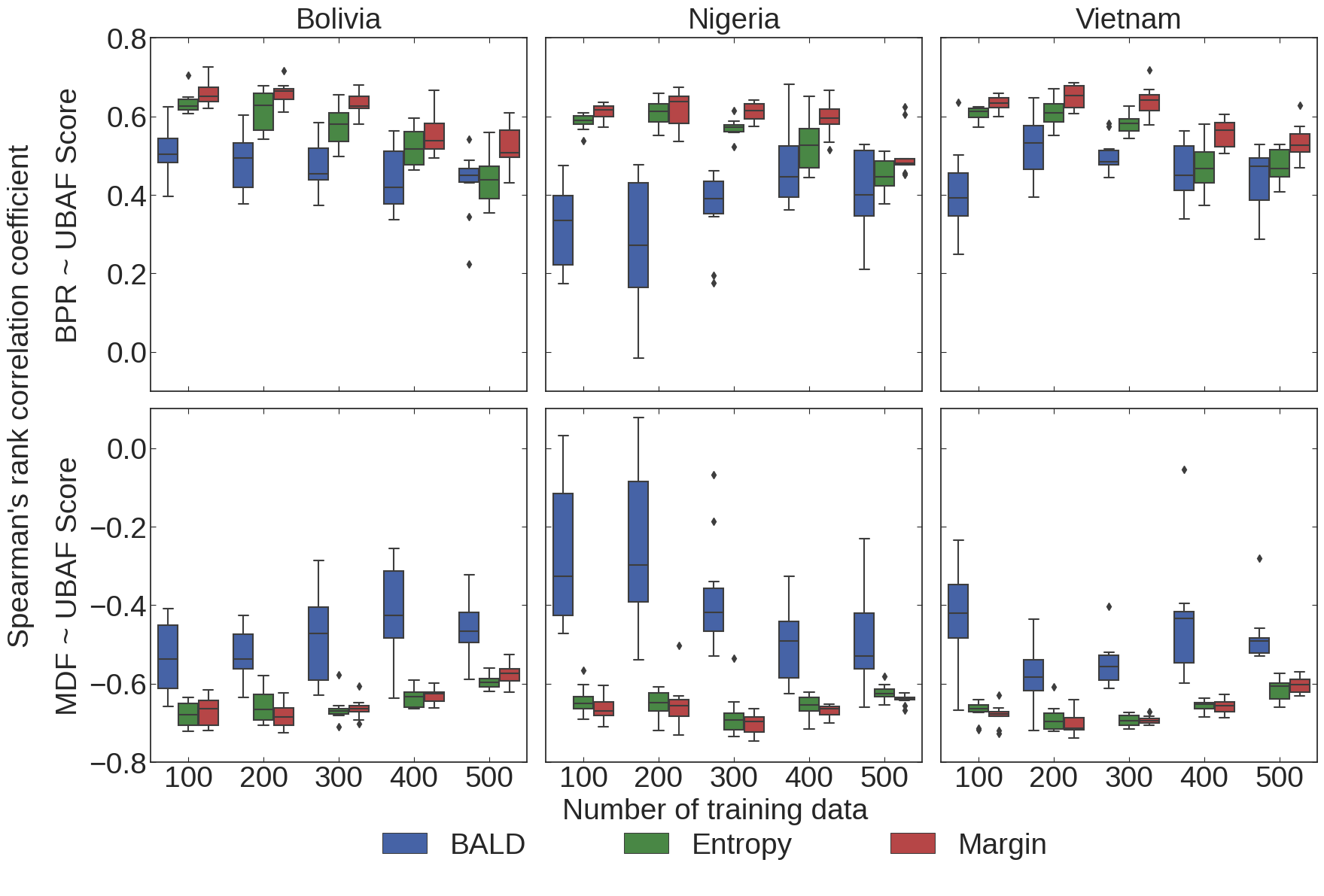}
  \caption{The comparison on the Spearman’s rank correlation coefficient between class ambiguity indices and the score of the uncertainty-based acquisition functions: (left) Bolivia, (middle) Nigeria, (right) Vietnam.}
  \label{fig:corr}
\end{figure}

In Fig. 8, we observed an evident positive rank correlation between the BPR, which reflects the ambiguity between classes at the tile level due to spatial resolution limitations in satellite imagery, and the score of the margin and entropy acquisition function, both of which are acquisition functions based on predictive uncertainty. In addition, across the three regions, the median correlation coefficients between the BPR and the margin acquisition function score were consistently the highest. In addition, the MDF, indicating tile-level ambiguity between classes due to spectral resolution limitations, exhibits a statistically significant negative correlation with the UAF score. Specifically, the scores from the margin and entropy acquisition function show a notable negative rank correlation with MDF. On the other hand, BALD, which represents model uncertainty, shows weaker rank correlation compared to the margin and entropy acquisition function, which utilize predictive uncertainty, in terms of the two class ambiguity indices.

When interpreting our findings through the uncertainty propagation theory, the experiment results suggest that class ambiguity arising from spatial and spectral resolution limitations of sensor in satellite imagery, as quantified by the BPR and MDF, significantly impacts the scores of acquisition functions based on predictive uncertainty, such as the margin and entropy acquisition functions. Consequently, by synthesizing the experiment results in Section 5.1 and 5.2, we draw the conclusion that the BPR and MDF are effective indicators to represent the informativeness of data points under the assumption of uncertainty-based acquisition function. 

\subsection{Visualization of Two-dimensional Density Plots Using the Class Ambiguity and Class Imbalance Indices}
\label{sec:Visualization of Two-dimensional Density Plots Using the Class Ambiguity and Class Imbalance Indices}

\subsubsection{The Distribution of Multi-spectral Satellite Images in the Unlabeled Data Pool }
\label{sec:The Distribution of Multispectral Satellite Images in the Unlabeled Data Pool}

Examining the distribution of data points within the unlabeled data pool is an important task for interpreting and understanding the behavior of the acquisition functions within the IDAL-FIM framework. Specifically, comparing the distribution of the unlabeled data pool with that of newly acquired data points selected by the acquisition functions helps clarify the behavior of the acquisition function. In this section, based on the established class ambiguity indices in Section 5.2, we employed two-dimensional (2D) density plots with the MDF on the x-axis and the BPR on the y-axis, named the MDF-BPR density plot, to visualize the distribution of data points in the IDAL-FIM framework. Furthermore, in order to interpret the behavior of the acquisition functions from the perspective of mitigating the class imbalance problem presented in the previous study \citep{ruuvzivcka2020deep}, we investigated the correlation between the FPR and the BPR in terms of spatial structure. Then, we visualized the distribution of data points in the unlabeled data pool by utilizing the FPR-BPR density plot which is the FPR on the x-axis and the BPR on the y-axis.

\begin{figure}[t]
  \centering
  \includegraphics[width=1.0 \linewidth]{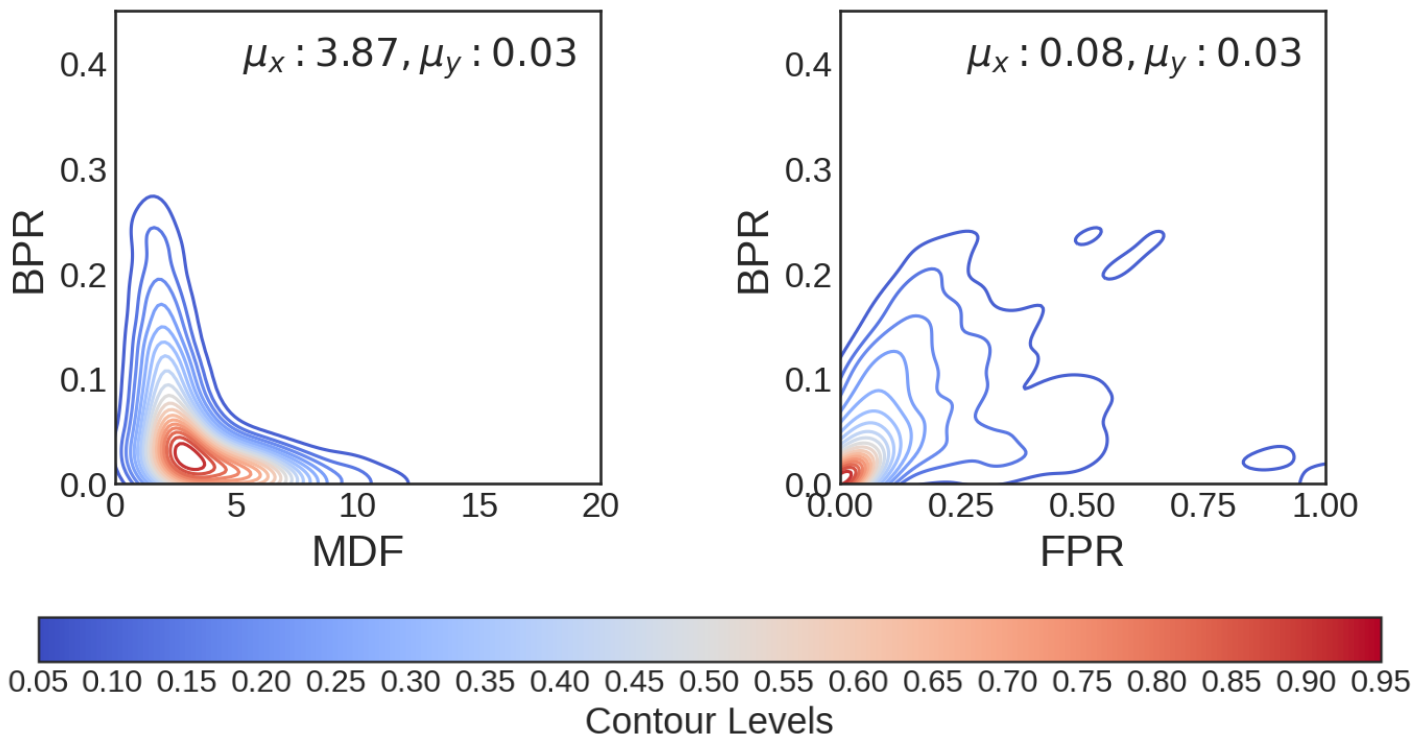}
  \caption{2D-density plots in the unlabeled data pool based on the class ambiguity and imbalance indices: (left) MDF-BPR density plot represents the data distribution in terms of class ambiguity. This plot shows the diversity of data points from a class ambiguity perspective; (right) FPR-BPR density plot displays the relationship between FPR and BPR. In each plot, $\mu_x$ denotes the average of x-axis values, and $\mu_y$ means the average of y-axis values.}
  \label{fig:dist-unlabeld-pool}
\end{figure}

Regarding the spatial structural relationship between the FPR and the BPR, we found a statistically significant correlation, indicating that BPR reaches its maximum value around an FPR of 0.5 in the given unlabeled dataset, using Pearson's correlation coefficient. Table 5 displays Pearson's correlation coefficient between the FPR and the BPR, categorized by the FPR threshold of 0.5 for the unlabeled data points. This result describes that when FPR is less than 0.5, FPR and BPR have a positive correlation, and otherwise, FPR and BPR have a negative correlation. 

\begin{table}[h]
\centering
\caption{Correlation coefficient between FPR and BPR in the unlabeled data pool}
\label{tab:Correlation coefficient between FPR and BPR in the unlabeled data pool}
\resizebox{0.8\textwidth}{!}{%
\begin{tabular}{cccc}
\hline
Criteria                & n    & Proportion (\%) & Pearson’s correlation coefficient \\ \hline
FPR \textless 0.5       & 1477 & 96.4            & 0.677                             \\
FPR \textgreater{}= 0.5 & 55   & 3.6             & -0.584    \\ \hline
\end{tabular}%
}
\end{table}

In both the MDF-BPR and the FPR-BPR plots, the contour lines were estimated using the kernel density function, and they represent levels that range from 0.05 to 0.95 at intervals of 0.05. These levels correspond to iso-proportions of the density. For instance, the contour line drawn for 0.05 is the outermost line on the two-dimensional density plot. This contour line represents the area where 5\% of the probability mass lies outside the contour lines. The same color of contour line represents an area at the same level. 
Fig. 9 displays the MDF-BPR and the FPR-BPR density plots of data points in the unlabeled data pool. In the MDF-BPR density plot, the average values of the x-axis (MDF) and y-axis (BPR) are 3.87 and 0.03, respectively, and the data points are widely distributed along both axes. This shape of distribution density means that the unlabeled data pool contains diverse data points from the perspective of the class ambiguity indices. Lower MDF values and higher BPR values indicate higher class ambiguity, whereas higher MDF values and lower BPR values can be interpreted as data points with lower class ambiguity. In the FPR-BPR density plot, the average values of the x-axis (FPR) and y-axis (BPR) are 0.08 and 0.03, respectively. This plot illustrates that the maximum value of BPR occurs when the FPR is 0.5, aligned with the result in Table 5. 

\subsubsection{The Distribution of Newly Acquired Multi-spectral Satellite Images within the IDAL-FIM framework}
\label{sec:The Distribution of Newly Acquired Multispectral Satellite Images within the IDAL-FIM framework}

In the same visualization manner as in Section 5.3.1, we depicted the distribution of the newly selected data points through acquisition functions in each iteration within the IDAL-FIM framework using MDF-BPR density plots and FPR-BPR density plots. Each figure consists of 20 2D-density plots representing five different acquisition functions during four iterations. The visualization results in Nigeria and Vietnam are similar to those in Bolivia; therefore, we present the Bolivia results as the representative visualization in Figs. 10 and 11. One notable observation in Figs. 10 and 11 is that the distribution of multi-spectral satellite images, selected by the acquisition function, is dependent on the distribution of the unlabeled data pool. In the case of the random acquisition function, the averages of the x- and y-axes remain similar over four iterations, and the shapes of the contour lines in the density plot resemble those found in the unlabeled pool, as depicted in Fig. 9. This observation apparently indicates that the multi-spectral satellite images selected by the random acquisition function are affected by the distribution of the unlabeled data pool. 

In addition, the K-means acquisition function, which is the density-based acquisition function utilized in this study, also exhibited similar shape of contour line patterns as the random acquisition function. Considering the assumption of the density-based acquisition function that data points maximizing the diversity of data features are informative, it can be inferred that this function is sensitive to the distribution of the unlabeled data pool. In addition, the experiment results support that K-means acquisition function is influenced by the distribution of the unlabeled data pool. Therefore, in scenarios where the distribution of the unlabeled data pool is not uniform, the effectiveness of the density-based acquisition function in selecting informative data points may be compromised.

On the other hand, the margin and entropy acquisition function, which are acquisition functions based on the predictive uncertainty, showed distinct shapes of contour line patterns. In Fig. 10, the margin and entropy acquisition function tend to select data points which have higher BPR values and lower MDF values compared to the random and K-means acquisition function. In particular, these characteristics were most evident especially in the first iteration out of the four iterations. This observation indicates that the margin and entropy acquisition function are capable of selecting data points with higher levels of class ambiguity in the input satellite image while minimizing dependence on the distribution of the unlabeled pool. Additionally, in Fig. 11, when the margin and entropy acquisition function select data points with higher BPR values, this leads to the selection of data points with FPR values around 0.5. This trend is particularly noticeable during the first iteration.

However, as the number of iterations increases, the margin and entropy acquisition functions progressively become influenced by the data distribution within the unlabeled data pool. This influence is illustrated by visualizing both the MDF-BPR and FPR-BPR density plots, which depict a pattern where the point representing the average values of each axis gradually moves toward the average of the corresponding density plot for the unlabeled data pool. Moreover, the shape of contour lines corresponding to the 95\% probability mass was the largest in the first of the four iterations and gradually decreased as the iterations progressed in both the MDF-BPR and the FPR-BPR density plots. 

\begin{figure}[p]
  \centering
  \includegraphics[width=1.0 \linewidth]{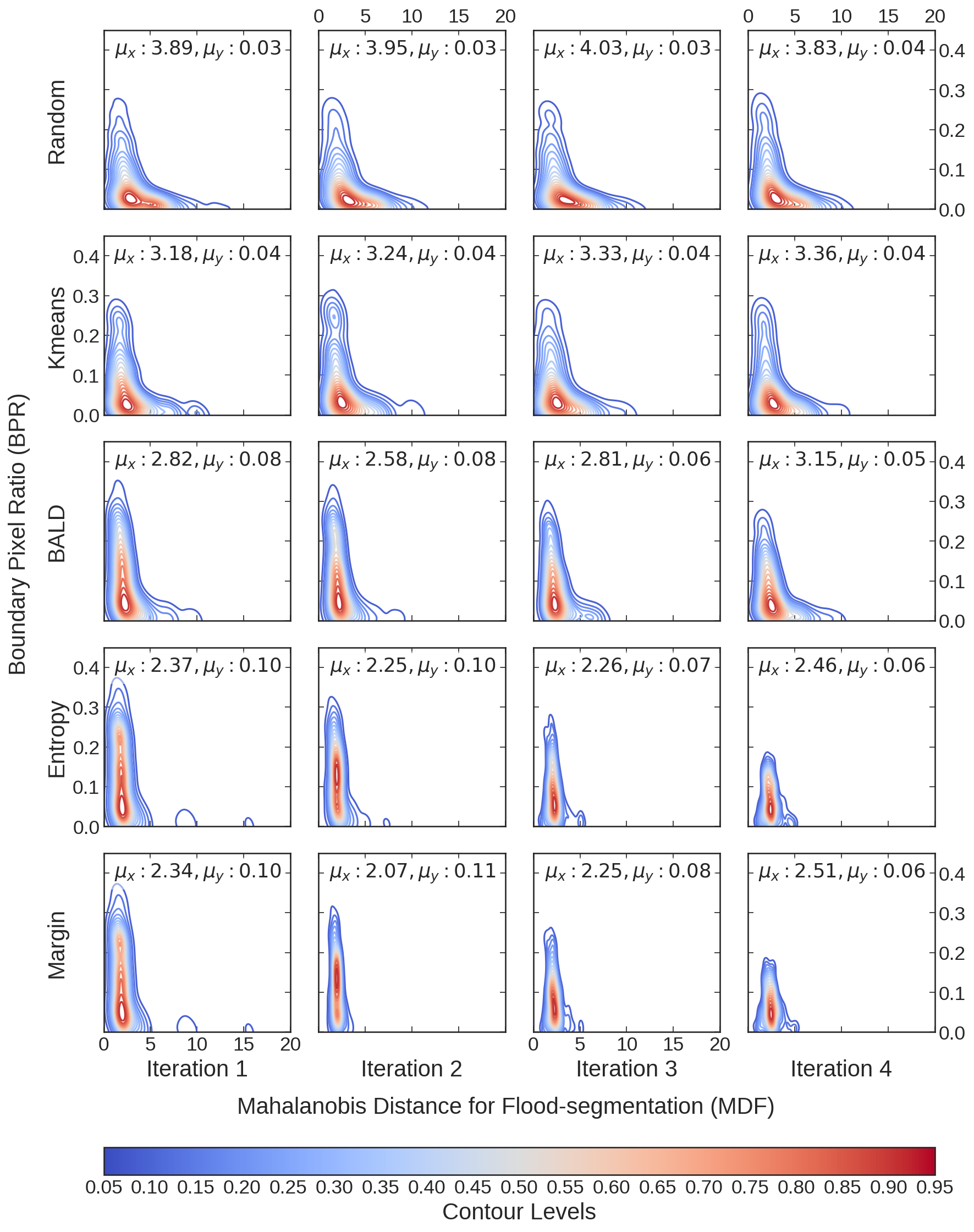}
  \caption{The MDF-BPR density plots in Bolivia. In each plot, $\mu_x$ denotes the average of x-axis values, and $\mu_y$ means the average of y-axis values.}
  \label{fig:bolivia-mdf-bpr}
\end{figure}

\begin{figure}[p]
  \centering
  \includegraphics[width=1.0 \linewidth]{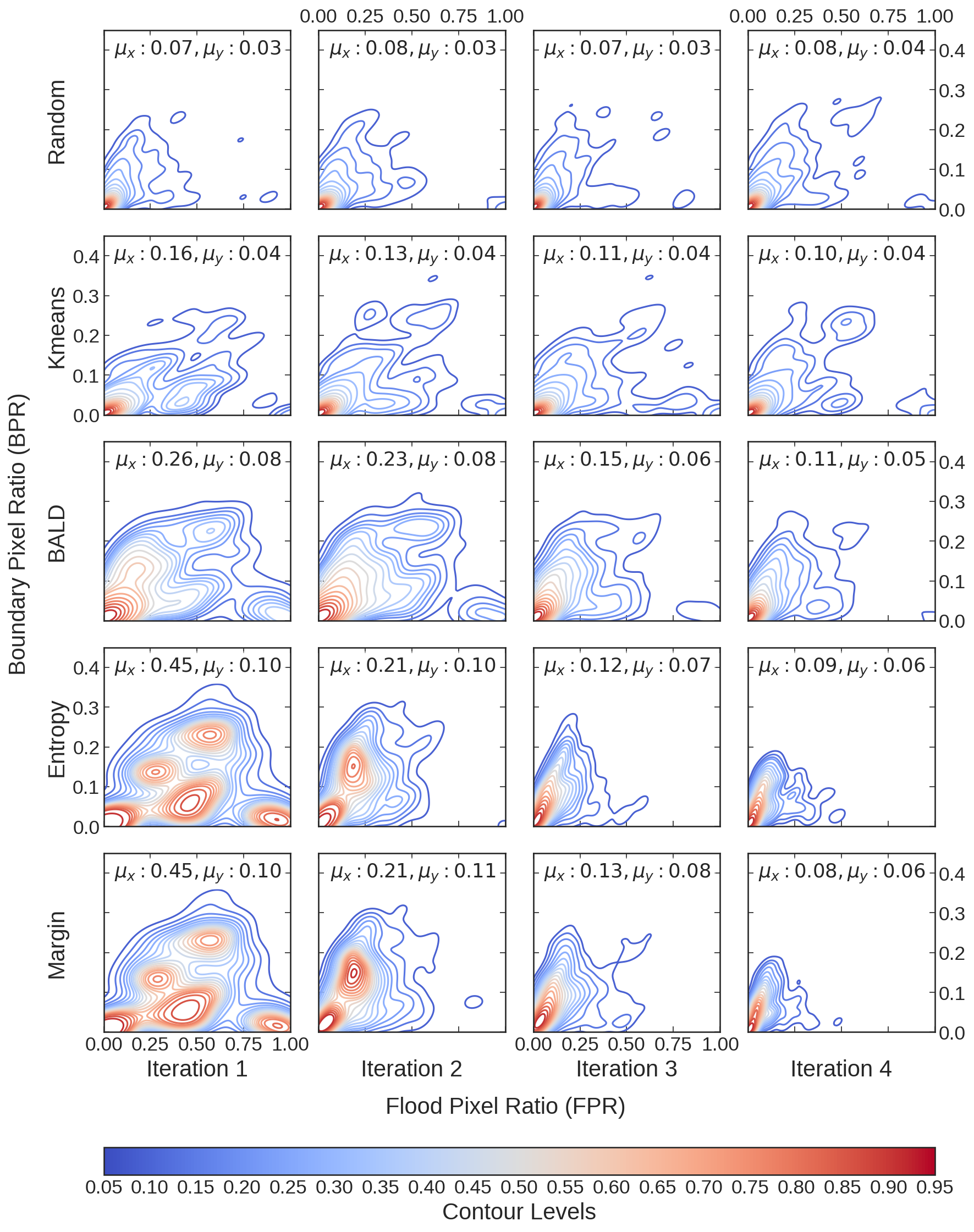}
  \caption{The FPR-BPR density plots in Bolivia. In each plot, $\mu_x$ denotes the average of x-axis values, and $\mu_y$ means the average of y-axis values.}
  \label{fig:bolivia-fpr-bpr}
\end{figure}

Lastly, the BALD acquisition function exhibited a different visual pattern than the other four acquisition functions. As depicted in Fig. 10, the BALD acquisition function demonstrated superior capability in selecting data points with high BPR compared to the random and K-means acquisition function, but it was not as proficient in identifying data points with low MDF compared to the margin and entropy acquisition function. Since BALD is a measure of model uncertainty, in situations where predictive uncertainty is high but consistent predictions are made, the score of the BALD acquisition function becomes low. Therefore, this property could cause the BALD acquisition function to struggle in identifying class ambiguity between flooded and non-flooded areas in multi-spectral satellite images at the tile level, especially when compared to the margin and entropy acquisition function.

\section{Discussion}
\label{sec:Discussion}

Through experiments in Section 5.1, we have shown that the margin acquisition function consistently achieves the best performance and the entropy acquisition function is the second-best performer within the IDAL-FIM framework. Additionally, in Section 5.2, we demonstrated that the two class ambiguity indices of input satellite images have statistically significant rank correlation with the score of the margin and entropy acquisition function. In Sections 5.1 and 5.2, both the margin and entropy acquisition function, which are based on the predictive uncertainty, showed comparable performance and correlation with class ambiguity indices. This can be explained by the fact that both are equivalent to querying the instance with a class posterior closest to 0.5 in the binary classification setting \citep{settles2009active}. Therefore, when combining our findings with the uncertainty propagation theory, the observed statistically significant correlations strongly support a causal relationship between the class ambiguity of the input satellite image and the score of the predictive uncertainty-based acquisition functions, such as the margin and entropy acquisition function. Consequently, we conclude that two class ambiguity indices, the BPR and MDF, are effective indicators to represent informative data points under the assumption of uncertainty-based acquisition functions. 

When comparing the margin and entropy acquisition functions with other acquisition functions, Fig. 10 illustrates a noticeable shift in the shape of the contour lines. This shift suggests that as the acquisition function becomes more capable of identifying informative data points, the remaining informative data points in the unlabeled pool are depleted more quickly. Consequently, it is crucial in practice to continually add more data points into the unlabeled pool to improve the diversity and representativeness of the training data. Our proposed class ambiguity indices can effectively monitor the condition of the unlabeled data pool, thereby facilitating an active learning strategy for flood mapping by assisting decision-making in its updates.

When considering the depletion of informative data points, the distribution patterns of the data points selected by different acquisition functions tend to be more effectively highlighted in early iterations. For this reason, we observed the most distinctive distribution of data points selected by acquisition functions in the first iteration of Fig. 10. In the first iteration of Fig. 10, the average BPR of the data points selected by the margin acquisition function is 0.1. This value is more than twice the average BPR value of the random acquisition function, which is 0.03. In the case of the MDF, the average value of the data points selected by the margin acquisition function is 2.34, on the other hand, the average value of data points selected through the random acquisition function is 3.89. 

The interpretations through MDF and BPR not only help in understanding how acquisition functions behave in terms of class ambiguity but also provide a summary of preliminary suggestions for data labeling similar to the previous study \citep{li2022suggestive}. Based on these findings, we suggest prioritizing the labeling of flood-observed satellite images where there are more boundary pixels between flood and non-flood classes, and where there is a small difference in average pixel values between those two classes. In addition, given that labeled data is necessary to select data points based on class ambiguity indices, the advantage of deep active learning becomes more evident, as informative data points can be chosen within statistical significance based on predictive uncertainty, even without labels.

Based on the spatial structural relationship between the BPR and the FPR, we demonstrate that the margin and entropy acquisition function have the capability to alleviate class imbalance issues. In Section 5.3.1, we showed that the BPR tends to reach a maximum when the FPR is 0.5. This means that as data points with high BPR are selected by the acquisition function, the FPR of selected data points tends to become concentrated around 0.5. In Fig. 11, continuing with the concepts of the informative data points depletion, during the first iteration, the margin and entropy acquisition function exhibits a distinctive preference for selecting higher BPR data points. This selection pattern is associated with the ability to select data points near an FPR of 0.5. Therefore, the first iteration of the margin and entropy acquisition function obviously illustrates its capability to select in mitigating class imbalance issues. This finding aligns with the conclusion in the previous study \citep{ruuvzivcka2020deep}, where the DAL framework was able to automatically balance classes in the training data, even when dealing with an extreme class imbalance in the pool of unlabeled data. \citet{ruuvzivcka2020deep} were able to maintain class balance over the iterations by selecting up to 950 pairs (1.1\%) of training data, which is smaller than the 1,072 pairs of “changed” class out of a total of 83,144 pairs. On the other hand, in our study, experiments were conducted using 500 samples (32.6\%) out of a total of 1,532 samples. As a result, since the proportion of selected data out of the total dataset is higher compared to the previous study \citep{ruuvzivcka2020deep}, the depletion of informative data points is displayed more evidently in Fig. 11, as iterations progress.

\section{Conclusion}
\label{sec:Conclusion}

In this paper, we introduced a novel framework of Interpretable Deep Active Learning for Flood inundation Mapping (IDAL-FIM) by leveraging class ambiguity indices based on the uncertainty propagation theory. In the experiments, we utilized Sen1Floods11 dataset, and adopted U-Net with MC-dropout as deep learning model for flood inundation mapping. We employed five acquisition functions, which are random, K-means, BALD, entropy, and margin acquisition function, within the IDAL-FIM framework. Based on the experiment results, we demonstrated the significance of two proposed class ambiguity indices within the IDAL-FIM framework. This is achieved by establishing their statistically significant correlation with the predictive uncertainty of the deep learning model at the tile level. Then, we illustrated that the behaviors of deep active learning are effectively interpreted using two class ambiguity indices within the IDAL-FIM framework, through visualizing two-dimensional density plots and providing explanations regarding the operation of deep active learning.

The limitations of this study are as follows. In flood mapping, one of the notable challenges is the distinction between flooded and non-flooded vegetated areas, as they generally exhibit comparable spectral patterns. Based on this study, satellite image tiles that observe both flooded and non-flooded vegetated areas are expected to be selected with high priority due to the high class ambiguity between those two classes. However, due to the absence of a distinct class for non-flooded vegetated areas in Sen1Floods11, we were not able to thoroughly explore the behavior of acquisition functions in terms of flooded and non-flooded vegetated areas. Therefore, further research is needed on the behavior of acquisition functions for such closely resembling classes within the IDAL-FIM framework. In addition, the labeling cost was only considered from the quantity point of view, and each individual labeling difficulty was not taken into account. Furthermore, as one of the characteristics of remote sensing data is multi-modality, multi-modal data, such as multi-spectral images, SAR (Synthetic Aperture Radar) images and DEM (Digital Elevation Model), are valuable datasets for flood inundation mapping. However, in this study, only multi-spectral images in the binary segmentation were considered for the interpretation of deep active learning. As a research direction for future studies, it is important to focus on an active learning framework that incorporates the multi-modality of remote sensing data and methods for their interpretation.



\bibliographystyle{elsarticle-harv} 
\bibliography{2023-DAL-FIM-reference}

\newpage

\listoffigures





\end{document}